\documentclass[sigconf]{acmart}

\usepackage{threeparttable}
\usepackage{dsfont}
\usepackage{amsmath}

\usepackage{amssymb}
\usepackage{diagbox}

\acmSubmissionID{1066}

\setcopyright{acmcopyright}
\copyrightyear{2021}
\acmYear{2021}
\setcopyright{acmcopyright}\acmConference[MM '21]{Proceedings of the 29th ACM International Conference on Multimedia}{October 20--24, 2021}{Virtual Event, China}
\acmBooktitle{Proceedings of the 29th ACM International Conference on Multimedia (MM '21), October 20--24, 2021, Virtual Event, China}
\acmPrice{15.00}
\acmDOI{10.1145/3474085.3475351}
\acmISBN{978-1-4503-8651-7/21/10}
\makeatletter
\def\authornotetext#1{
 \g@addto@macro\@authornotes{%
 \stepcounter{footnote}\footnotetext{#1}}%
}
\makeatother

\begin{document}
\fancyhead{}

\title{Disentangle Your Dense Object Detector}


\author[Zehui Chen, Chenhongyi Yang, Qiaofei Li, Zheng-Jun Zha, Feng Zhao, and Feng Wu]{
	Zehui Chen\textsuperscript{\rm 1,}*, 
	Chenhongyi Yang\textsuperscript{\rm 2,}*, 
	Qiaofei Li\textsuperscript{\rm 3}, 
	Feng Zhao\textsuperscript{\rm 1,\dag},
	Zheng-Jun Zha\textsuperscript{\rm 1},
	Feng Wu\textsuperscript{\rm 1}
}
\authornotetext{Equal contribution}
\authornotetext{Corresponding author}
\def\authors{Zehui Chen, Chenhongyi Yang, Qiaofei Li, Zheng-Jun Zha, Feng Zhao, and Feng Wu}

\affiliation{
	\textsuperscript{\rm 1} University of Science and Technology of China, Hefei
	\country{China}\\
	\textsuperscript{\rm 2} University of Edinburgh, Edinburgh
	\country{United Kingdom}\\
	\textsuperscript{\rm 3} SenseTime, Shanghai
	\country{China}
}

\email{lovesnow@mail.ustc.edu.cn, chenhongyi.yang@ed.ac.uk, }
\email{liqiaofei1@senseauto.com, {fzhao956, zhazj, fengwu}@ustc.edu.cn}
\renewcommand{\shortauthors}{}
\begin{abstract}
    Deep learning-based dense object detectors have achieved great success in the past few years and have been applied to numerous multimedia applications such as video understanding. However, the current training pipeline for dense detectors is compromised to lots of conjunctions that may not hold. In this paper, we investigate three such important conjunctions: 1) only samples assigned as positive in classification head are used to train the regression head; 2) classification and regression share the same input feature and computational fields defined by the parallel head architecture; and 3) samples distributed in different feature pyramid layers are treated equally when computing the loss. We first carry out a series of pilot experiments to show disentangling such conjunctions can lead to persistent performance improvement. Then, based on these findings, we propose Disentangled Dense Object Detector (DDOD), in which simple and effective disentanglement mechanisms are designed and integrated into the current state-of-the-art dense object detectors. Extensive experiments on MS COCO benchmark show that our approach can lead to 2.0~mAP, 2.4~mAP and 2.2~mAP absolute improvements on RetinaNet, FCOS, and ATSS baselines with negligible extra overhead. Notably, our best model reaches 55.0 mAP on the COCO \textit{test-dev} set and 93.5 AP on the hard subset of WIDER FACE, achieving new state-of-the-art performance on these two competitive benchmarks. Code is available at \href{https://github.com/zehuichen123/DDOD}{https://github.com/zehuichen123/DDOD}. 
\end{abstract}



\maketitle

\section{Introduction}


With the recent advance of deep learning, visual object detection has achieved massive progress in both performance and speed~\cite{bochkovskiy2020yolov4,wang2020scaled,lin2017focal}. It has become the foundation for widespread multimedia applications, such as animation  editing and interactive video games. Compared with the pioneering two-stage object detectors~\cite{ren2015faster,cai2018cascade}, single-stage dense detectors have received increasing attentions because of their simple and elegant pipeline. It directly outputs the detected bounding boxes, which ensures a fast inference speed, making it more suitable to be deployed on edge-computing devices. 

\begin{figure}[!t]
    \includegraphics[width=\columnwidth]{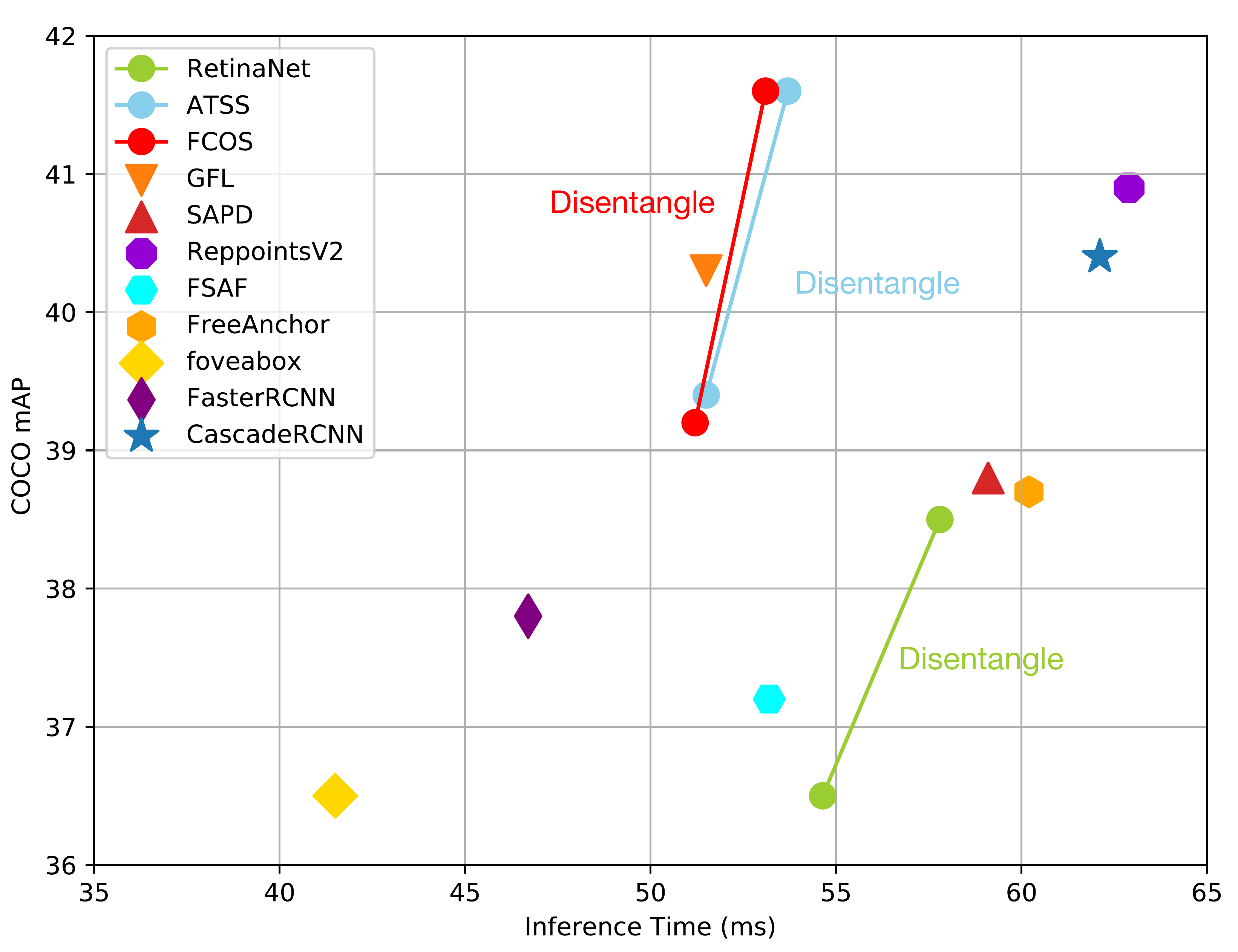}
    \caption{Our DDOD can improve the performance of different state-of-the-art dense object detectors with negligible extra inference time.}
    \label{fig:fig1}
    \vspace{-0.3cm}
  \end{figure}

A typical dense object detector is composed of three parts: 1) a backbone network, usually pretrained on ImageNet~\cite{deng2009imagenet}, that extracts useful features from the input image and outputs multi-scale feature maps to detect objects at various scales; 2) a feature pyramid network (FPN)~\cite{lin2017feature} that further improves the multi-scale feature maps through feature fusion, in which shadow features are promoted by high-level semantic features from deeper layers; 3) a detection head that usually contains two branches to output the corresponding classification and bounding box regression results at each anchor point in the input feature map.

Since the first dense detector DenseBox~\cite{huang2015densebox} is proposed, hundreds of work has enhanced the performance, both in accuracy and speed. For example, Focal loss~\cite{lin2017focal} is one of the most prominent methods. It lowers down the gradient magnitude of well-classified samples, hence forcing the model to concentrate on hard samples. Recently, several anchor-free detectors have been proposed to remove the labor-intense anchor designing procedure. Other techniques~\cite{wang2020end,sun2020onenet} focus on label assignment where the classic IoU threshold is replaced with more meticulous strategies. In a recent approach~\cite{jiang2018acquisition}, a localization quality estimation branch is added to the detection head to help with keeping high-quality boxes during non-maximum-suppression (NMS). 

Although many algorithms have been proposed to improve the performance of dense detectors, the room for further improvement is still large. Specifically, we notice that there are several conjunctions in the training process of dense detectors that can lead to performance deterioration. These conjunctions are usually the compromises of engineering implementations, which seem reasonable but have not been studied thoroughly before.

\textbf{Firstly}, the regression loss is only applied to the positive samples of classification branch. This design sounds correct because we only care about the localization of foreground objects. However, will other samples benefit the training of the regression head?  

\textbf{Secondly}, regardless of what the final bounding box looks like, the receptive fields, defined by several consecutive convolution operations, for classification and regression tasks are exactly the same. However, as pointed out in~\cite{song2020revisiting}, these two tasks have different preferences: classification desires regions with rich semantic information while regression prefers to attend edge parts. Therefore, the same receptive fields cannot guarantee the optimal performance. 

\textbf{Finally}, there is an imbalance problem lying in the pyramid layers because of the same supervision on all training samples. Concretely, the number of training samples on $P_3$ level is 256 ($2^8$) times to that on $P_7$. Thus, concatenating them together to compute the training loss will result in extremely imbalanced supervision among samples on different layers, as the shallow layers receives much more supervision than deeper layers like $P_7$. 

To summarize, the aforementioned conjunctions can be divided into three types: 1) the label assignment conjunction that the regression branch is only trained using the positive samples in classification branch; 2) the spatial feature conjunction between classification and regression where the same receptive field makes the model attend the same area of the feature map; 3) the pyramid supervision conjunction between different layers of FPN, in which the results on different layers are flattened and concatenated for loss computation, leading to the same supervision of all the layers. 

In this paper, we first carry out a series of meticulously designed pilot experiments to show that disentangling such conjunctions can significantly improve the overall detection performance. Then, we propose a new training pipeline for dense object detectors called \textbf{DDOD} (\textbf{D}isentangled \textbf{D}ense \textbf{O}bject \textbf{D}etector) with excellent performance, as illustrated in Figure~\ref{fig:fig1}. Specifically, for the assignment conjunction, we design separated label assigners for classification and regression, enabling us to pick out the most suitable training sets for those two branches, respectively. As for the spatial feature conjunction, an adaptive feature disentanglement module based on deformable convolution is proposed to automatically attend different features that benefit classification and regression. Finally, for the pyramid supervision conjunction, we design a novel re-weighting mechanism to adaptively adjust the magnitude of supervision on different FPN layers based on the positive samples on each layer. 

The main contributions of this work are three-fold:
\begin{itemize}
\item We carry out detailed experiments to investigate three types of training conjunctions in dense object detectors, which are usually ignored by previous work but can cause performance deterioration.
\item We propose to decouple these conjunctions in a unified detection framework called \textbf{DDOD}, which consists of label assignment disentanglement, spatial feature alignment disentanglement and pyramid supervision disentanglement. 
\item Through extensive experiments, we validate the effectiveness of the proposed DDOD on two competitive dense object detectors and achieve state-of-the-art performance on both COCO and WIDER FACE datasets. 
\end{itemize}

\section{Related Work}

\subsection{Label Assignment in Object Detection}
Selecting which anchors are to be designated as positive or negative labels has been viewed as a crucial task, affecting the detector performance greatly. A vanilla practice is to assign anchors whose IoUs with the ground truth are greater than a certain threshold~\cite{lin2017focal, ren2015faster}. With the advance of anchor-free models, a new strategy for defining labels was proposed by FCOS~\cite{tian2019fcos}, which directly views points inside GT boxes as positives. FreeAnchor~\cite{zhang2021learning} constructs a bag of anchor candidates for each GT and solves the label assignment as a problem of detection-customized likelihood optimization. ATSS~\cite{zhang2020bridging} suggests an adaptive anchor assignment which is calculated by the statistics from the mean and standard deviation of IoU values on a set of anchors for each GT. In order to adaptively separate anchors according to the models's learning status, PAA~\cite{kim2020probabilistic} investigates a probabilistic manner to assign labels. However, these methods do not discuss the label conjunction between classification and regression branches.
\subsection{Feature Alignment in Object Detection}
Better feature alignment always indicates better performance. CascadeRPN ~\cite{NEURIPS2019_d554f7bb}, GuidedAnchor ~\cite{wang2019region} and AlignDet ~\cite{chen2019revisiting} adopt a cascade-manner to improve the feature representation via re-aligning features. By re-extracting features at the afore-regressed bounding boxes, one can predict more precise classification scores and regression boxes. Besides, There are a variety of approaches that discuss the feature representation conjunction in classification and regression tasks on two-stage detectors. DoubleHead~\cite{wu2020rethinking} is the first one to find this phenomenon and proposes separate heads with different structures for classification and regression heads. TSD~\cite{song2020revisiting} improves the "shared head for classification and localization"~(sibling head), firstly demonstrated in Fast RCNN by adopting different {\tt RoIAlign} offsets for classification and regression. Although this conjunction has been validated and well-addressed in two-stage methods, whether this problem exist in dense detectors remains an open problem.
\subsection{Imbalanced Learning in Object Detection}
Imbalanced learning exists in various aspects for many object detection frameworks~\cite{pang2019libra,oksuz2020imbalance}. Alleviating such unbalance issues are crucial to attain satisfying detection results. RetinaNet~\cite{lin2017focal} proposes Focal loss to conquer extreme imbalance problem between the number of samples in positive and negative, which enables us to train dense object detectors without sampling strategy. Pang et al.~\cite{pang2019libra} noticed that the detection performance is often limited by the imbalance during the training process, which generally consists in three aspects, namely the sample level, feature level and objective level. To address the above issues, Libra-RCNN is proposed to mitigate such gaps with IoU-balanced sampling, balanced feature pyramid, and balanced L$_{1}$ loss. Besides, the category imbalance is also a vital topic in object detection~\cite{tan2020equalization,tan2020equalizationv2,wu2020forest}. EQL~\cite{tan2020equalization} finds the accumulation of backward gradients from negative samples is larger than those from positive samples in rare categories, which results in the insufficient training on infrequent classes. Therefore, an equalization loss was proposed to moderate such a phenomenon. \\

In comparison, our work aims to disentangle these three important conjunctions in state-of-the-art dense object detectors to achieve better accuracy with minimum cost. 

\section{Pilot Experiments}
In this section, we present our motivation by carrying out pilot experiments to show the harm of three conjunctions in dense object detectors and the possibility to advance the detection performance by disentanglement.
\subsection{Label Assignment Conjunction}
\label{sec:3.1}
The assignment conjunction refers to the case that the regression loss is only applied to the "foreground" samples in classification task. Actually, this conjunction is here for historical reasons. The concept of bounding box regression was first introduced in~\cite{girshick2014rich}, in which a smooth L$_1$ loss is used to train the regression branch. Applying the regression loss only to positive samples, whose IoU with the GT box is larger than a threshold, ensures the network to get enough information from the feature map for estimating the actual location of the object. However, this threshold does not need to coordinate with the classification threshold, and there is no such "positive" concept when inferring as every anchor box or center point will be applied with its predicted regressor. Thus, setting a separate threshold for bounding box regression is reasonable. To validate our assumption, we train different RetinaNet models with various classification and regression IoU thresholds ranging from 0.4 to 0.6 and report the results in Table~\ref{tab:pilot-encode}. It can be seen that the optimal performance, with a mAP of 36.5, is achieved with an IoU of 0.5 on classification head and 0.4 on regression head. Therefore, in our DDOD, we decompose the assignment conjunction by adopting respective label assigners on classification and regression branches for better detection accuracy.

\begin{table}[!hbpt]
	\caption{AP performance with different IoU thresholds for classification and regression branches on label assignment.}
	\centering
		\begin{threeparttable}
			\begin{tabular}{c| c c c c}
				\toprule
				\diagbox{${IoU}_{cls}$}{${IoU}_{reg}$} & 0.4 & 0.5 & 0.6 \\
				\midrule
				0.4 & 35.0 & 35.3 & 33.2 \\
				0.5 & \textbf{36.5} & 36.1 & 35.3 \\
				0.6 & 34.6 & 35.2 & 35.1\\
				\bottomrule
			\end{tabular}
		\end{threeparttable}
		\label{tab:pilot-encode}
\end{table}

\subsection{Spatial Feature Conjunction}
\begin{figure}[!t]
    \includegraphics[width=0.9\columnwidth]{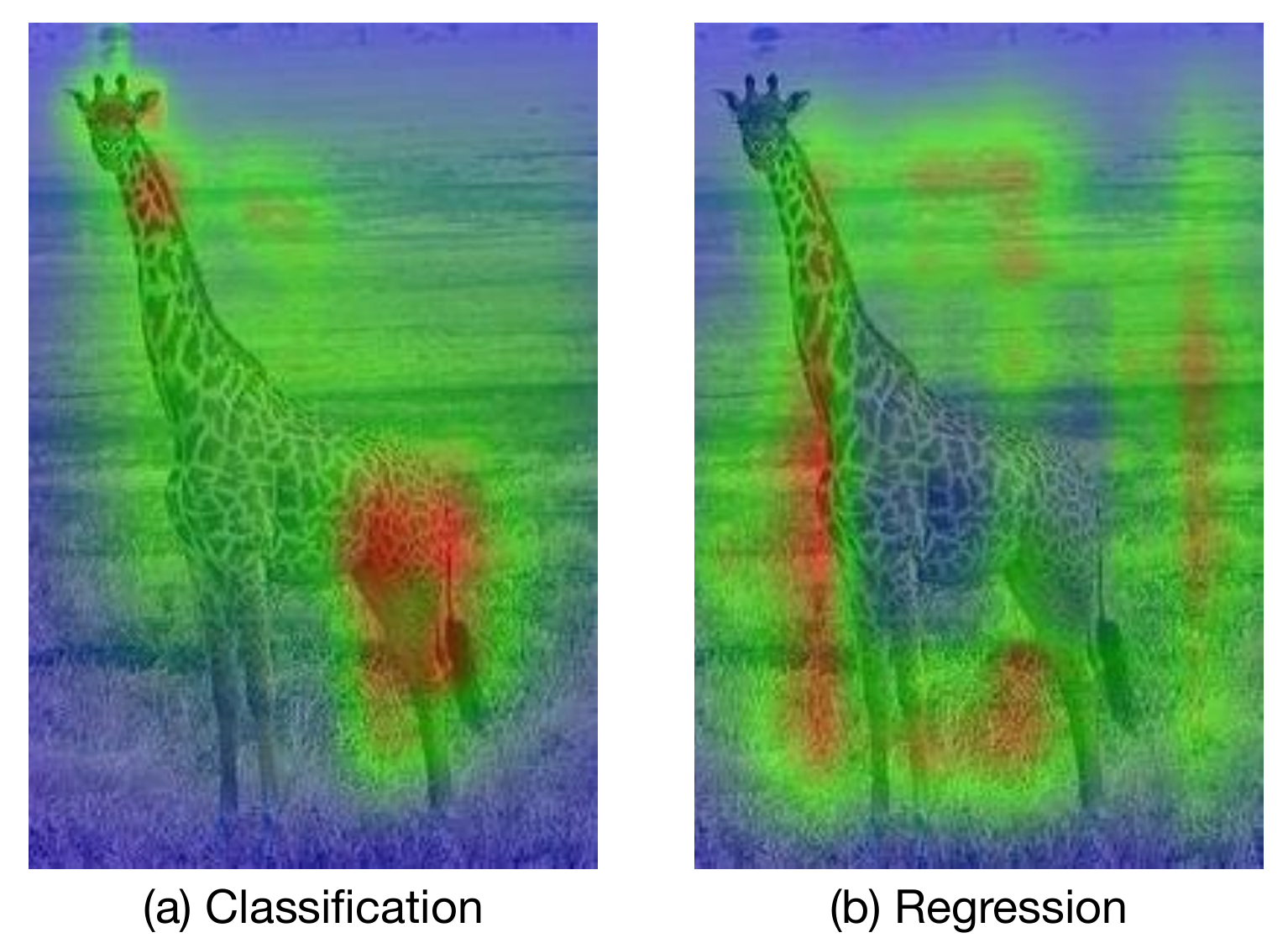}
    \caption{Visualization of backward gradient on the output of FPN. The heatmap is obtained by utilizing GradCAM~\cite{zhou2018interpretable}. }
    \label{fig:fpn_cam}
    \vspace{-0.3cm}
  \end{figure}

In a typical dense object detector, two parallel branches with the same structure are used for classification and regression, which implies that their corresponding receptive areas in the feature maps are the same. However, as pointed in TSD~\cite{song2020revisiting}, in two stage detectors these two tasks should concentrate on different parts of the objects: regions that is rich of semantic information is helpful for classification while the contour areas are essential for localization. We argue that this principle also holds for dense detectors. In Figure~\ref{fig:fpn_cam}, we visualize the sensitive regions using GradCAM~\cite{zhou2018interpretable} for classification and regression and find that they are indeed different, which validates the necessity to disentangle their receptive regions. In Section~\ref{sec:4.2}, we will describe a simple design to achieve this goal by adding an adaptive feature disentanglement module in both branches.

\subsection{Supervision Conjunction in FPN}
\label{sec:3.3}
In the commonly adopted implementation of feature pyramid-based dense detectors, the predictions of different layers are "flatten and concatenated" for loss computation, making the supervision on different layers to be conjectural. However, as pointed out in~\cite{yang2021querydet}, the sample distribution is significantly imbalanced between different layers, because the feature size grows up in quadratic as the feature resolution increases. As a result, the training samples will be dominated by samples in lower levels, making the high-level samples lack of supervision, which will harm the performance on large objects. A simple idea to overcome this is to assign a larger weight to high-level samples to compensate for the supervision. In Table~\ref{tab:pilot-fpn}, we exploit different re-weighting strategies, where only the weights for lowest and highest layers are set and the rest ones are linearly interpolated. The best strategy is to linearly increase the weight from 1.0 to 2.0 and the worst one is to linearly decrease the weight from 3.0 to 1.0. This result validates our assumption and inspires us to propose an effective re-weighting approach called FPN hierarchical loss to disentangle the supervision conjunction.

\begin{table}[h]
	\caption{AP performance with linear increasing (decreasing) supervision weights from $P_3$ to $P_7$ levels on ATSS baseline.}
	\centering
		\begin{threeparttable}
			\begin{tabular}{c| c | c c | c c c}
				\toprule
				re-weighting factor & AP &AP$_{50}$ &AP$_{75}$ &AP$_{\it S}$ &AP$_{\it M}$ &AP$_{\it L}$\\
				\midrule
				1$\sim$1 & 39.4 & 56.6 & 42.6 & \textbf{23.9} & 42.5 & 49.6\\
				1$\sim$2 & \textbf{39.8} & \textbf{57.8} & \textbf{42.9} & 22.9 & \textbf{43.5} & 51.8\\
				1$\sim$3 & 39.4 & 56.7 & 42.9 & 21.6 & 43.1 & \textbf{52.0}\\
				2$\sim$1 & 38.7 & 56.6 & 42.1 & 23.2 & 42.7 & 48.3\\
				3$\sim$1 & 37.6 & 55.3 & 40.8 & 22.2 & 41.6 & 47.0\\
				\bottomrule
			\end{tabular}
		\end{threeparttable}
		\label{tab:pilot-fpn}
\end{table}

\section{Methodologies}
In this section, we describe the proposed DDOD in detail. An overview of our approach is presented in Figure~\ref{fig:method_vis}.

\subsection{Label Assignment Disentanglement}
\label{sec:4.1}
One of the most widely accepted practice in label assignment is to only regress samples that are assigned as positives in the classification branch. However, as exploited in Section~\ref{sec:3.1}, such a conjunction between the two branches is suboptimal. Motivated by this observation, we design different label assignment strategies for classification and regression, respectively. In the early work, IoU between the candidate anchor and the ground truth is the only criterion for foreground-background assignment. Recently, some good attempts show that replacing IoU with the loss value is a more suitable approach for label assignment. Thus, we adopt a clean and effective cost formulation proposed in \cite{wang2020end} to mine more appropriate samples for different tasks. Given an image $I$, suppose there are $P$ predictions based on the predefined anchors and $N$ ground truth. For each candidate (anchor or center point) $P_{i}$, the foreground probability $\hat{p}(i)$ and the regressed bounding box $\hat{b}(i)$ is output with respect to each category. To this end, a \textit{Cost Matrix} can be formulated as:
\begin{figure*}[!hbpt]
    \includegraphics[width=\linewidth]{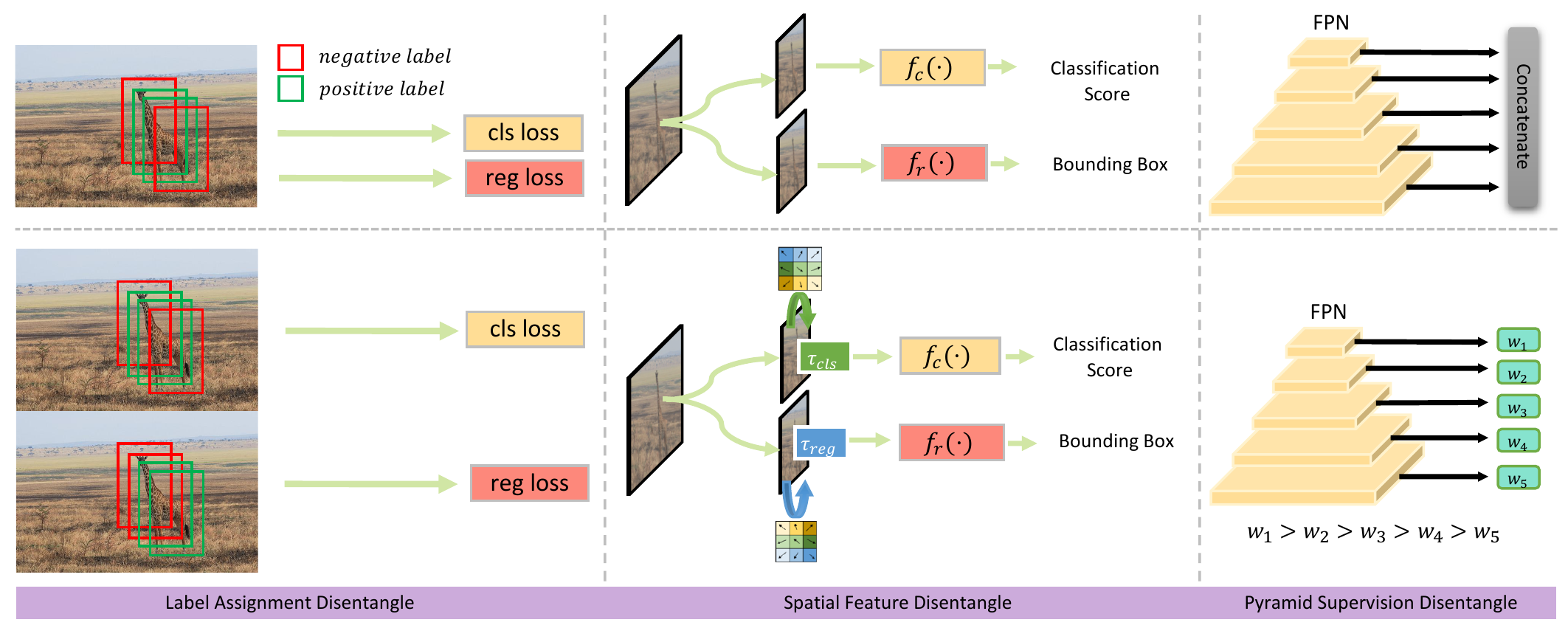}
	\caption{The proposed DDOD consists of 3 parts: label assignment disentanglement, spatial feature disentanglement, and pyramid supervision disentanglement. For each component, our approach (bottom) is compared with the original implementation (top). For label assignment, our method yields different pos/neg labels for the giraffe in classification and regression branches. For spatial feature disentanglement, we propose to learn separate convolutional offset for each branch that aims at providing spatial aligned features for each head. The supervision disentanglement tackles the imbalanced learning problem with different loss weight for each FPN layer where the original one treated all the samples equally.}
    \label{fig:method_vis}
    \vspace{-0.3cm}
  \end{figure*}
  
\begin{equation}
\begin{small}
	\begin{aligned}
        C_{i,\pi(i)} = \underbrace{\mathds{1} \left[ \pi(i) \in \Omega_i \right]}_{\text{spatial prior}} ~\cdot~ \underbrace{{\Bigl( \hat{p}_{\pi(i)}(i) \Bigr)}^{1-\alpha}}_{\text{classification}} ~\cdot \underbrace{{\Bigl( \mathrm{IoU} \bigl( b_i, \hat{b}_{\pi(i)}(i) \bigr) \Bigr)}^\alpha,}_{\text{regression}}
    \label{eq:quality}
    \end{aligned}
\end{small}
\end{equation}
where $C_{i,\pi(i)} \in [0, 1]$ represents the matching quality of the $\pi(i)$-th prediction with respect to each ground truth, and $\Omega_i$ denotes the set of candidate predictions for $i$-th ground truth. To stabilize the training process and make model converge faster, only candidates whose centers fall into the ground-truth boxes are considered as possible foreground samples. In the assignment process, top $K$ predictions with the largest cost value are selected from each FPN level. Then, the candidates are assigned as foreground samples if their matching quality is beyond an adaptive threshold computed with the batch statistics like ATSS.

In order to decouple the label assignment of classification and regression, we introduce a hyper-parameter $\alpha \in [0, 1]$ to balance the contribution between them. Intuitively, samples with higher foreground probability should be prioritized for classification, while the regression branch focuses more on regression quality. A detailed ablation study on $\alpha$ will be elaborated in Section~\ref{sec:5.4.1}. In doing so, classification and regression tasks are disentangled successfully, through which both objectives are achieved in a better way. 

\subsection{Spatial Feature Disentanglement}
\label{sec:4.2}
Our pilot experiments have shown the difference between the most informative features for classification and regression. Here, we describe a simple but effective approach to spatially disentangle these features with the help of deformable convolution. Similar ideas~\cite{song2020revisiting,wu2020rethinking} have been proved successful in two-stage detectors, and here we extend them to dense detectors. The gist is to let the model attend the most important feature for decision-making, but such a disentanglement is somewhat conflict with the parallel head architecture in state-of-the-art dense object detectors. However, with the help of deformable convolution, the model is able to attend different spatial fields in the input feature map using the same number of convolutions. Inspired by this, we introduce the \textit{adaptive feature disentanglement module}, in which a learnable convolution offset is predicted on each position $p_n$ to adaptively select the spatial features for the current task. Formally, for each position $p_i$ on the feature map $F$, we have
\begin{align}
		F(p_i) = \sum_{p_n \in \textit{R}} w(p_n) \cdot x(p_i + p_n + \Delta p_n).
\end{align}
Instead of sampling using a regular grid $R$ over the input feature map, we let the model learn the best fitted position $\Delta p_n$ to convolute for each task. In this paradigm, the classification branch focuses more on class-wise discriminative features, e.g., the human face of a person or the wheel of a car. On the other hand, the regression branch pays more attention to border features. Since the offset should be correlated with the information at current position, the offset $\Delta p_n$ is predicted with additional filters. In the actual implementation, this operation is implemented through deformable convolution~\cite{dai2017deformable}. We directly replace the first convolution with deformable convolution in the  classification and regression subnets. Hence, features from FPN can be dynamically extracted given the different tasks at the same time, without competing with each other. 
\subsection{Pyramid Supervision Disentanglement}
\label{sec:4.3}
Samples located at different levels of feature pyramid are entangled together with equal supervision since the FPN~\cite{lin2017feature} was proposed. However, as described in Section~\ref{sec:3.3}, this practice does not consider the imbalance contribution of training samples from different layers. Concretely, objects located at $P_3$ level receive more attention because of the huge number of training samples on it. On the contrary, objects fall into $P_7$ level are less likely to be well-trained due to the lack of training samples. A natural solution to alleviate this phenomenon is to strengthen the supervision in deeper layers. In practice, this can be easily achieved through assigning samples in deeper layers with larger weights. For this purpose, we propose \textit{FPN hierarchical loss} on top of the original supervision. The reason behind the gradient imbalance issue is the number of samples falling in different layers varies. Therefore, we directly choose the count of samples as the metric to indicate whether we should down-weight/up-weight supervision during training. The re-weighting coefficient $w$ for the sample located on layer $L_i$ with respect to the loss $L$ is formulated as: 
\begin{align}
		w_{cls} &= 2 - \frac{n_{cls}^{i} - min(N_{cls})}{max(N_{cls}) - min(N_{cls})} \\
		w_{reg} &= 2 - \frac{n_{reg}^{i} - min(N_{reg})}{max(N_{reg}) - min(N_{reg})}\\
		L &= w_{cls} \cdot L_{cls} + w_{reg} \cdot L_{reg},
		\label{eq:alpha_compute}
\end{align}
where $n_{cls}^i$ and $n_{reg}^i$ are the number of positive samples located at FPN level $i$ based on the aforementioned classification and regression branches, $N_{cls}$ and $N_{reg}$ are the set of $n_{cls}^i$ and $n_{reg}^i$ respectively. When the re-weighting factor $w$ ranges from 1.0 to 2.0, we empirically find that samples on $P_3$ level always get a factor of 1.0 and those on $P_7$ level are always 2.0. From the gradient balance perspective, such results are beneficial for large object learning. Besides, we adopt a moving average manner to relieve the distributional difference of labels among images and stabilize the training process. More specifically, we use the accumulated number of positive samples $\sum_{j=0}^{J}{n^i_{cls/reg}}$ at each FPN level $i$ until current iteration $J$ to compute $w$.

\section{Experiments}
\begin{table}[t]
	\begin{threeparttable}
	\centering
	\caption{Comparison of detection AP results on different single-stage object detectors (RetinaNet, FCOS, and ATSS).}
	\begin{tabular}{c|c|c|cc|ccc}
		\toprule
		Model & DDOD & AP &AP$_{50}$ &AP$_{75}$ &AP$_{\it S}$ &AP$_{\it M}$ &AP$_{\it L}$\\
		\midrule
		RetinaNet & & 36.5 & 55.4 & 39.1 & 20.4 & 40.3 & 48.1\\
		RetinaNet & \checkmark & 38.5 & 58.1 & 41.2 & 20.7 & 41.4 & 51.6\\
		\midrule
		FCOS \tnote{*} & & 39.2 & 57.3 & 42.4 & 22.7 & 43.1 & 51.5\\
		FCOS \tnote{*} & \checkmark & 41.6 & 59.7 & 45.3 & 24.3 & 45.0 & 55.1\\
		\midrule
		ATSS & & 39.4 & 56.6 & 42.6 & 23.9 & 42.5 & 49.6\\
		ATSS & \checkmark & 41.6 & 59.9 & 45.2 & 23.9 & 44.9 & 54.4\\
		\bottomrule
	\end{tabular}
	\begin{tablenotes}
		\small
		\item[*] The advanced version of FCOS implementation in ATSS is adopted.
	\end{tablenotes}
	\label{tab:diff_detectors}
	\end{threeparttable}
\end{table}
\begin{table*}
	\begin{threeparttable}
	\centering
	\caption{Effects of each component in our DDOD. Results are reported on COCO \textit{val-2017} with ATSS.}
	\begin{tabular}{ccc|c|cc|ccc}
		\toprule
		Assignment Disentangle & Spatial Feature Disentangle & Supervision Disentangle & AP &AP$_{50}$ &AP$_{75}$ &AP$_{\it S}$ &AP$_{\it M}$ &AP$_{\it L}$\\
		\midrule
		 & & & 39.4 & 56.6 & 42.6 & 23.9 & 42.5 & 49.6\\
		\checkmark & & & 40.4 & 58.9 & 43.8 & \textbf{24.1} & 43.9 & 51.4\\
		 & \checkmark & & 40.6 & 58.6 & 44.1 & 24.0 & 44.1 & 52.5\\
		 & & \checkmark & 40.0 & 57.9 & 43.1 & 23.0 & 43.6 & 52.6\\
		 \checkmark & \checkmark & \checkmark & \textbf{41.6} & \textbf{59.9} & \textbf{45.2} & 23.9 & \textbf{44.9} & \textbf{54.4}\\
		\bottomrule
	\end{tabular}
	\label{tab:ablation}
	\end{threeparttable}
\end{table*} 

In this section, we construct extensive experiments on COCO~\cite{lin2014microsoft} and WIDER FACE~\cite{yang2016wider} datasets to study the proposed DDOD. We first present the effectiveness and generality of our method. Secondly, we show the ablation results of different modules. Thirdly, we conduct experiments to discuss some interesting aspects to gain deeper insight of the proposed DDOD. Finally, we compare our approach with other state-of-the-art models and show our superiority.

\begin{table}[!h]
	\caption{AP performance with different $\alpha$ for classification and regression label assignment strategies.}
	\centering
		\begin{threeparttable}
			\begin{tabular}{c| c c c c}
				\toprule
				\diagbox{$\alpha_{cls}$}{$\alpha_{reg}$} & 0.5 & 0.6 & 0.7 & 0.8 \\
				\midrule
				0.5 & 39.4 & 39.3 & 39.0 & 39.0\\
				0.6 & 39.6 & 39.4 & 39.5 & 39.5\\
				0.7 & 39.9 & 40.1 & 39.7 & 39.6\\
				0.8 & \textbf{40.4} & 40.0 & 40.0 & 39.9\\
				\bottomrule
			\end{tabular}
		\end{threeparttable}
	\label{tab:ablation_alphas}
\end{table}

\subsection{Implementation Details}
\label{sec:5.1}
Most experiments are carried out on COCO benchmark, where we train the model on \textit{trainval35k} with 115K images and report performance on \textit{minval} with 5K images. Besides, we report the final results on \textit{test-dev} with 20K images from the evaluation server. Except for the results on \textit{test-dev}, all experiments are trained under standard 1x schedule setting based on ResNet-50~\cite{he2016deep}, with default parameters adopted in MMDetection~\cite{chen2019mmdetection}. Moreover, in order to adapt cost function-based label assignment, we replace the centerness branch with IoU-prediction branch. In addition to COCO, we also validate our approach on WIDER FACE, which is currently the largest face detection dataset. The model is trained with a batch size of 24 on 6 Titan V100s. The schedule of learning rate is annealing down from 7.5e-3 to 7.5e-5 every 30 epochs using cosine decay rule. This process is repeated for 20 times, indicating a total epoch number of 600. In order to align our baseline with other face detectors, deformable convolution block, SSH module, and DIoU loss are adopted. 

\subsection{Results on Dense Object Detectors}
We implement DDOD on three different representative dense detectors: RetinaNet, FCOS, and ATSS. The mAP performance is reported in Table~\ref{tab:diff_detectors}. Our DDOD boosts all the baselines by more than 2.0~mAP, which validates the effectiveness and generation ability of the proposed method. When observing the results in detail, we find that the AP$_L$ is promoted most in all three detectors. The reasons can be two-fold. Firstly, large objects are more prominent in the image plain, so our spatial disentangle module is easier to attend important parts. 
Secondly, Our FPN hierarchical loss is proposed to compensate for the insufficient supervision for high level feature maps, so the mAP for large objects is directly benefited.

\begin{table}[!hbpt]
	\caption{AP performance with different positions of deformable convolution in classification and regression heads.}
	\centering
		\begin{threeparttable}
			\begin{tabular}{c| c |c c| c c c}
				\toprule
				Position & AP &AP$_{50}$ &AP$_{75}$ &AP$_{\it S}$ &AP$_{\it M}$ &AP$_{\it L}$\\
				\midrule
				1 & \textbf{40.6} & 58.4 & \textbf{44.3} & 23.6 & \textbf{44.3} & 52.4\\
				2 & 40.3 & \textbf{58.5} & 44.1 & \textbf{23.9} & 44.0 & 52.4\\
				3 & 40.1 & 57.9 & 43.4 & 23.0 & 44.6 & 52.4\\
				4 & 40.1 & 57.9 & 43.1 & 23.0 & 44.6 & \textbf{52.6}\\
				\bottomrule
			\end{tabular}
		\end{threeparttable}
		\label{tab:dcn-pos}
\end{table}
\begin{table}[!hbpt]
	\caption{AP performance with various re-weighting strategies for FPN hierarchical loss.}
	\centering
		\begin{threeparttable}
			\begin{tabular}{c| c |c c| c c c}
				\toprule
				Method & AP &AP$_{50}$ &AP$_{75}$ &AP$_{\it S}$ &AP$_{\it M}$ &AP$_{\it L}$\\
				\midrule
				Sum-Reweight & 39.7 & 57.5 & 43.2 & 22.6 & 43.3 & 52.1\\
				Sum-Reweight\tnote{*} & 39.8 & 57.8 & 43.1 & 23.0 & 43.3 & 51.0\\
				Linear-Reweight & 39.8 & 57.8 & 42.9 & 22.9 & 43.5 & 51.8\\
				Linear-Reweight\tnote{*} & 39.8 & 57.7 & 42.9 & 23.0 & 43.5 & 52.0\\
				Linear-Interpolate & 39.8 & 57.9 & 43.1 & \textbf{23.1} & 43.2 & 51.1\\
				Linear-Interpolate\tnote{*} & \textbf{40.0} & \textbf{57.9} & \textbf{43.4} & 22.9 & \textbf{43.6} & \textbf{52.6}\\
				\bottomrule
			\end{tabular}
			\begin{tablenotes}
				\small
				\item[*] indicates the implementation of moving average version by counting all valid samples till current iteration.
			\end{tablenotes}
		\end{threeparttable}
		\label{tab:reweight}
\end{table}

\subsection{Ablation Study} 
To understand how each disentanglement in DDOD facilitates detection performance, we test each component independently on the baseline detector ATSS and report its AP performance in Table~\ref{tab:ablation}. The overall AP of the baseline starts from 39.4. When label assignment disentanglement is applied, the AP is raised by 1.0 and the improvement is seen on objects of all sizes. This result vindicates our findings in Section~\ref{sec:4.1} that decoupling the encoding conjunctions between classification and bounding box regression is necessary. Then, we add the spatial feature disentanglement that brings us an 1.2 AP enhancement. Specifically, the AP$_{L}$ is benefited most, which validates our first assumption in Section~\ref{sec:4.2}. Another interesting finding is that the improvement on AP$_{75}$ brought by spatial feature disentanglement achieves 1.5, suggesting the underlying connection between correctly attending receptive regions and accurate bounding box localization. When pyramid supervision disentanglement is added, the accuracy is promoted by 0.6 and the AP$_L$ is raised by 3.0. This large improvement is in correspondence with our analysis in Section~\ref{sec:4.3} and the underlying intuition of this disentanglement. Finally, the AP achieves 41.6 when all three disentanglement are applied, gaining an 2.2 absolute improvement and validating our approach.


\begin{table*}
	\renewcommand\arraystretch{1.2}
	\centering
	\footnotesize \setlength{\tabcolsep}{4.5pt}
	\begin{threeparttable}
		\resizebox{\textwidth}{!}{
			\begin{tabular}{l|l|c|c|c|ccc|ccc|c}
				\hline
				Method  & Backbone & Epoch & MS$_{\text{train}}$ & FPS & AP &AP$_{50}$ &AP$_{75}$ &AP$_{\it S}$ &AP$_{\it M}$ &AP$_{\it L}$ & Reference\\
				\hline
				\textit{multi-stage:}  & & & & & & & & & & & \\
				Faster R-CNN w/ FPN \cite{lin2017feature}  & R-101 & 24 & & 14.2 & 36.2 & 59.1 & 39.0 & 18.2 & 39.0 & 48.2 & CVPR17 \\
				Cascade R-CNN \cite{cai2018cascade}  & R-101 & 18 & & 11.9 &42.8 &62.1 &46.3 &23.7 &45.5 &55.2 & CVPR18\\
				Grid R-CNN \cite{lu2019grid} & R-101 & 20 &  & 11.4 &41.5 & 60.9 & 44.5 & 23.3 & 44.9 & 53.1 & CVPR19 \\
				Libra R-CNN \cite{pang2019libra} & X-101-64x4d & 12 & & 8.5 & 43.0 & 64.0 & 47.0 & 25.3 & 45.6 & 54.6 & CVPR19\\
				RepPoints \cite{yang2019reppoints}   &R-101 & 24 & & 13.3 & 41.0 & 62.9 & 44.3 & 23.6 & 44.1 & 51.7 & ICCV19\\
				RepPoints \cite{yang2019reppoints}   &R-101-DCN& 24 & $\checkmark$ & 11.8 & 
				45.0 & 66.1 & 49.0 & 26.6 & 48.6 & 57.5 & ICCV19 \\
				RepPointsV2 \cite{chen2020reppoints} & R-101 & 24 & $\checkmark$ & 11.1 & 46.0 & 65.3 & 49.5 & 27.4 & 48.9 & 57.3 & NeurIPS20 \\
				RepPointsV2 \cite{chen2020reppoints} & R-101-DCN & 24 & $\checkmark$ & 10.0 &  48.1 & 67.5 & 51.8 & 28.7 & 50.9 & 60.8 & NeurIPS20 \\
				TridentNet \cite{li2019scale}& R-101 & 24 & $\checkmark$ & 2.7 & 42.7 & 63.6 & 46.5 & 23.9 & 46.6 & 56.6 & ICCV19 \\
				TridentNet \cite{li2019scale}& R-101-DCN & 36 & $\checkmark$ & 1.3 & 46.8 & 67.6 & 51.5 & 28.0 & 51.2 & 60.5 & ICCV19 \\
				TSD \cite{song2020revisiting} & R-101 & 20 &  & 1.1 & 43.2 & 64.0 & 46.9 & 24.0 & 46.3 & 55.8 & CVPR20\\
				BorderDet \cite{qiu2020borderdet} & R-101 & 24 & $\checkmark$ & 13.2 &  45.4 & 64.1 & 48.8 & 26.7 & 48.3 & 56.5 & ECCV20\\
				BorderDet \cite{qiu2020borderdet} & X-101-64x4d & 24 & $\checkmark$ & 8.1 &  47.2 & 66.1 & 51.0 & 28.1 & 50.2 & 59.9 & ECCV20\\
				BorderDet \cite{qiu2020borderdet} & X-101-64x4d-DCN & 24 & $\checkmark$ & 6.4 &  48.0 & 67.1 & 52.1 & 29.4 & 50.7 & 60.5 & ECCV20\\
				\hline
				\hline
				\textit{one-stage:}   & & & & & & & & & & & \\
				TPNet320 \cite{10.1145/3394171.3413691} & R-101 & 24 & & 25.7 & 34.2 & 53.1 & 36.4 & 13.6 & 36.8 & 50.5 & ACM MM20\\
				TPNet512 \cite{10.1145/3394171.3413691} & R-101 & 24 & & 13.9 & 39.6 & 58.5 & 42.8 & 20.5 & 45.3 & 53.3 & ACM MM20\\
				CornerNet \cite{law2018cornernet} &HG-104 & 200 & $\checkmark$ & 3.1 & 40.6 & 56.4 & 43.2 & 19.1 & 42.8 & 54.3 & ECCV18 \\
				CenterNet \cite{duan2019centernet} &HG-104 & 190 &  $\checkmark$ & 3.3 &44.9 &62.4 &48.1 &25.6 &47.4 &57.4 & ICCV19 \\
				CentripetalNet \cite{dong2020centripetalnet} & HG-104 & 210 & $\checkmark$ & n/a  & 45.8 & 63.0 & 49.3 & 25.0 & 48.2 & 58.7 & CVPR20\\
				RetinaNet \cite{lin2017focal} &R-101 &  18 &  & 13.6 &39.1 &59.1 &42.3 &21.8 &42.7 &50.2 & ICCV17 \\
				FreeAnchor \cite{NEURIPS2019_43ec517d} & R-101 & 24 & $\checkmark$ & 12.8 & 43.1 & 62.2 & 46.4 & 24.5 & 46.1 & 54.8 & NeurIPS19 \\
				FreeAnchor \cite{NEURIPS2019_43ec517d} & X-101-32x8d & 24 & $\checkmark$ & 8.2 & 44.9 & 64.3 & 48.5 & 26.8 & 48.3 & 55.9 & NeurIPS19 \\
				FSAF \cite{zhu2019feature} & R-101 & 18 & $\checkmark$ & 15.1 & 40.9 & 61.5 & 44.0 & 24.0 & 44.2 & 51.3 & CVPR19\\
				FSAF \cite{zhu2019feature}& X-101-64x4d & 18 & $\checkmark$ & 9.1 & 42.9 & 63.8 & 46.3 & 26.6 & 46.2 & 52.7 & CVPR19 \\
				FCOS \cite{tian2019fcos} & R-101 & 24 & $\checkmark$ & 14.7 &41.5 & 60.7 & 45.0 & 24.4 & 44.8 & 51.6 & ICCV19\\
				FCOS \cite{tian2019fcos} & X-101-64x4d & 24 & $\checkmark$ & 8.9 & 44.7 & 64.1 & 48.4 & 27.6 & 47.5 & 55.6 & ICCV19\\
				SAPD \cite{zhu2019soft}& R-101 & 24 & $\checkmark$ & 13.2 & 43.5 & 63.6 & 46.5 & 24.9 & 46.8 & 54.6 & CVPR20 \\
				SAPD \cite{zhu2019soft}&  R-101-DCN & 24 & $\checkmark$ & 11.1 & 46.0 & 65.9 & 49.6 & 26.3 & 49.2 & 59.6 & CVPR20 \\
				SAPD \cite{zhu2019soft}& X-101-32x4d-DCN & 24 & $\checkmark$ & 8.8 & 46.6 & 66.6 & 50.0 & 27.3 & 49.7 & 60.7 & CVPR20 \\
				ATSS \cite{zhang2020bridging}& R-101 & 24 & $\checkmark$ & 14.6 &43.6 &62.1 &47.4 &26.1 &47.0 &53.6 & CVPR20 \\
				ATSS \cite{zhang2020bridging} &R-101-DCN & 24 &  $\checkmark$ & 12.7 & 46.3 &64.7 &50.4 &27.7 &49.8 &58.4 & CVPR20\\
				ATSS \cite{zhang2020bridging}& X-101-32x8d-DCN & 24 & $\checkmark$ & 6.9 & {47.7} & {66.6} &{52.1} & {29.3} &50.8 & {59.7} & CVPR20 \\
				PAA \cite{kim2020probabilistic} & R-101 & 24 & $\checkmark$ & 14.6 & 44.8 & 63.3 & 48.7 & 26.5 & 48.8 & 56.3 & ECCV20\\
				PAA \cite{kim2020probabilistic} & R-101-DCN & 24 & $\checkmark$ & 12.7 & 47.4 & 65.7 & 51.6 & 27.9 & 51.3 & 60.6 & ECCV20\\
				PAA \cite{kim2020probabilistic} & X-101-64x4d-DCN & 24 & $\checkmark$ & 6.9 & 49.0 & 67.8 & 53.3 & 30.2 & 52.8 & 62.2 & ECCV20\\
				GFL \cite{li2020generalized} &R-50 & 24 & $\checkmark$ & 19.4 & 43.1 & 62.0 & 46.8 & 26.0 & 46.7 & 52.3 & NeurIPS20\\
				GFL \cite{li2020generalized}&R-101 & 24 & $\checkmark$ & 14.6 & 45.0 & 63.7 & 48.9 & 27.2 & 48.8 & 54.5 & NeurIPS20 \\
				GFL \cite{li2020generalized}&R-101-DCN & 24 & $\checkmark$ & 12.7 & 47.3 & 66.3 & 51.4 & 28.0 & {51.1} & 59.2& NeurIPS20 \\
				GFL \cite{li2020generalized}&X-101-32x4d-DCN & 24 &  $\checkmark$ & 10.7 & {48.2} & {67.4} & {52.6} & {29.2} & {51.7} & {60.2}& NeurIPS20 \\
				\hline
				DDOD \textbf{(ours)}&R-50 & 24 & $\checkmark$ & 18.6 & 45.0 & 63.5 & 49.3 &  27.2 & 47.9 & 55.9 & -- \\
				DDOD \textbf{(ours)}&R-101 & 24 & $\checkmark$ & 14.0 & 46.7 & 65.3 & 51.1 & 28.2 & 49.9 & 57.9 & -- \\
				DDOD \textbf{(ours)}&R-101-DCN & 24 & $\checkmark$ & 12.0 & 49.0 & 67.7 & 53.6 & 29.6 & 52.2 & 62.3 & -- \\
				DDOD \textbf{(ours)}&X-101-32x4d-DCN & 24 &  $\checkmark$ & 10.3 & 50.1 & 69.0 & 54.8 & 30.8 & 53.5 & 63.2 & -- \\
				DDOD \textbf{(ours)}&R2-101-DCN & 24 &  $\checkmark$ & 10.4 & 51.2 & 69.9 & 55.9 & 32.2 & 54.4 & 64.9 & -- \\
				DDOD-X \textbf{(ours)}&R2-101-DCN & 24 &  $\checkmark$ & 8.4 & 52.5 & 71.0 & 57.2 & 33.9 & 56.8 & 65.9 & -- \\
				DDOD-X \textbf{(ours)} + MS$_{\text{test}}$ &R2-101-DCN & 24 &  $\checkmark$ & - & \textbf{55.0} & \textbf{72.6} & \textbf{60.0} & \textbf{37.0} & \textbf{58.0} & \textbf{66.7} & -- \\
				\hline
			\end{tabular}
		}
	\end{threeparttable}
    \captionsetup{font={small}}
	\caption{Comparisons between state-of-the-art detectors on COCO {\tt test-dev} \emph{(single-model and single-scale results except for the last row)}. ``MS$_{\text{train}}$'' and ``MS$_{\text{test}}$'' denote multi-scale training and testing, respectively; FPS values are measured on a single NVIDIA Titan V100 machine, using a batch size of 1 whenever possible; ``n/a'' means that both trained models and timing results from original papers are not available; \textbf{R}: ResNet \cite{he2016deep}; \textbf{X}: ResNeXt \cite{xie2017aggregated}; \textbf{HG}: Hourglass \cite{newell2016stacked}; \textbf{DCN}: Deformable Convolutional Network \cite{dai2017deformable}; \textbf{R2}: Res2Net \cite{gao2019res2net}.}
	\vspace{-6pt}
	\label{tab:coco_big}
\end{table*}

\begin{table}[!ht]
    \caption{AP performance of different methods on WIDER FACE \textit{val} subset. Our DDOD-Face achieves state-of-the-art results on AP hard under the same single-model test setting.}
     \centering
       \begin{threeparttable}
         \begin{tabular}{ c| c| c c c}
           \toprule
           Method & Backbone & Easy & Medium & Hard \\
           \cmidrule(r){1-1} \cmidrule(r){2-2} \cmidrule(r){3-5}
           PyramidBox \cite{tang2018pyramidbox} & VGG-16 & 0.961 & 0.950 & 0.889 \\
           PyramidBox++ \cite{li2019pyramidbox++} & VGG-16 & 0.967 & 0.959 & 0.912 \\
           AInnoFace \cite{zhang2019accurate} & ResNet-152 & 0.970 & 0.961 & 0.918 \\
           RetinaFace \cite{deng2019retinaface} & ResNet-152 & 0.969 & 0.961 & 0.918 \\
           RefineFace \cite{zhang2020refineface} & ResNet-152 & 0.972 & 0.962 & 0.920 \\
           DSFD \cite{li2019dsfd} & ResNet-152 & 0.966 & 0.957 & 0.904\\
           ASFD-D6 \cite{zhang2020asfd} & ResNet-152 & \textbf{0.972} & \textbf{0.965} & 0.925\\
           HAMBox \cite{Liu_2020_CVPR} & ResNet-50 & 0.970 & 0.964 & 0.933\\
           TinaFace \cite{zhu2020tinaface} & ResNet-50 & 0.970 & 0.963 & 0.934 \\
           \cmidrule(r){1-1} \cmidrule(r){2-2} \cmidrule(r){3-5}
           ATSS & ResNet-50 & 0.964 & 0.958 & 0.930 \\
           DDOD-Face \textbf{(ours)} & ResNet-50 & 0.970 & 0.964 & \textbf{0.935}\\
           \bottomrule
         \end{tabular}
       \end{threeparttable}
     \label{tab:widerface}
   \end{table}

\subsection{Discussion}
In this section, we delve into our DDOD to study how the mAP improvements are actually achieved and to gain deeper understanding of the underlying mechanisms. For all experiments, we employ ResNet-50-based ATSS with the same settings in Section~\ref{sec:5.1}.

\subsubsection{What is the Best Label Assignment Disentanglement?} 
\label{sec:5.4.1}
We conduct experiments to investigate the optimal strategy for our encoding disentanglement, i.e., to find the best $\alpha$ for classification and regression respectively. As presented in Table~\ref{tab:ablation_alphas}, the results are quite straight-forward: for regression branch, $\alpha = 0.5$ works better, while $\alpha = 0.8$ works best for classification. Although one extra hyper-parameter is introduced for label assignment, we find this setting ($\alpha_{cls} = 0.8, \alpha_{reg} = 0.5$) is robust to other scenarios when applying it on WIDER FACE dataset. The difference between optimal $\alpha$ values validates that the classification and regression branches should have their own label assignment criteria. We also observe that increasing $\alpha_{cls}$ will benefit the accuracy, suggesting that the IoU between anchors and ground-truth boxes are decisive for setting classification target. On the other hand, decreasing $\alpha_{reg}$ is good for the AP performance, indicating that the regression encoding should take the real-time classification results into consideration.

\subsubsection{Optimal Spatial Feature Disentanglement under Given Computational Budget.} 
Empirically, replacing some convolution operations with deformable convolution can somewhat bring performance enhancement, but seeking the best way for such replacement is nontrivial, as we want to avoid unbearable extra computational cost and the optimization difficulty brought by to many DCN layers. Hence, we only replace one convolution with DCN in each branch, and study the optimal way for the replacement. Concretely, we carry out experiments by replacing each convolution with its deformable version independently and check the final performance. From Table~\ref{tab:dcn-pos}, we find that disentangling feature representation at the first stage works best, with 40.6~mAP on COCO validation set. An interesting phenomenon is that the overall performance decrease when moving feature disentanglement to the latter layer in detection heads, indicating that disentangling features directly from FPN eases the learning process.

\subsubsection{Seeking the Most Robust Re-weighting Strategy.} 
In this part, we examine different re-weighting schemes to show that our proposed linear-interpolation re-weighting works best. The ablation results are shown in Table~\ref{tab:reweight}. Firstly, we test the Linear-Reweight proposed in~\cite{yang2021querydet}, in which the factors that linearly increase from 1.0 to 2.0 are assigned to each FPN level. Secondly, considering that most negative samples are greatly suppressed by the focal loss, we propose a vanilla re-weight strategy where the weight of each FPN layer is proportionate to the number of positive samples, which is denoted as Sum-Reweight. Additionally, since the number of positive samples can vary greatly according to the input batch, we extend the Linear-Reweight, Sum-Reweight, and our proposed Linear-Interpolate to that the weights are moving-averaged during optimization. This extension can stabilize the training process and slightly improve the accuracy according to Table~\ref{tab:reweight}. From the results, we can conclude that our linear-interpolation beats the other two policies, achieving 40.0 mAP on COCO \textit{val} subset.

\subsection{Comparison with State-of-the-Arts} 
As shown in Table~\ref{tab:coco_big}, we compare DDOD with state-of-the-art methods on COCO~\textit{test-dev}. Following the common practice, we adopt multi-scale training to improve model robustness. For fair comparison, results with single-model and single-scale testing at a short side of 800 are reported. When using the same backbone network, our DDOD beats all other methods including the recently developed GFL~\cite{li2020generalized} and PAA~\cite{kim2020probabilistic}. Additionally, empowered by the advanced Res2Net-101-DCN backbone, DDOD achieves a suppressing 52.5 mAP in the single-scale testing protocol, suggesting its great prospect in various applications. Finally, we incorporate techniques like multi-scale testing, Soft-NMS and stronger data augmentation for further performance improvement, achieving 55.0 mAP, which establishes new state-of-the-art performance on this competitive benchmark. 

\subsection{Results on WIDER FACE Dataset}
In addition to general object detection, our method is also applicable in other scenarios like face detection. We evaluate DDOD on the WIDER FACE benchmark, the currently largest open-sourced face detection dataset. It provides many difficult and realistic scenes with the varieties of occlusion, scale, and illumination. In those complex and crowded scenes, our model is quite robust and even outperforms several detectors designed for face detection, as shown in Table~\ref{tab:widerface}. In order to get better performance on WIDER FACE, which is different from COCO, we adopt DCN~\cite{dai2017deformable}, SSH module~\cite{najibi2017ssh}, and DIoU loss~\cite{zheng2020distance} on original ATSS model, reaching an overall AP of 93.0. Though this result surpasses several competitive face detectors, DDOD is still able to improve it by approximately 0.5~AP under easy, medium and, hard settings. More importantly, our DDOD-Face with ResNet-50 backbone achieves 97.0\%, 96.4\%, 93.5\% in these three settings on the validation subset, reaching state-of-the-art result on AP hard under the same single-model test setting. 

\section{Conclusion}
In this work, we develop the DDOD, a new training paradigm for dense object detectors. DDOD decomposes conjunctions lying in most current one-stage detectors via label assignment disentanglement, spatial feature disentanglement, and pyramid supervision disentanglement. The findings in this paper can potentially be harnessed to improve the detection accuracy and applied on various dense object detectors. We have demonstrated that DDOD achieves state-of-the-art results on COCO and WIDER FACE benchmarks, and validated its effectiveness and generalization as well. Furthermore, our approach may help to improve the performance of two-stage detectors. Therefore, how to embed DDOD into RCNN-like detectors will be an important next step to generalize DDOD.

\clearpage
{
\bibliographystyle{ACM-Reference-Format}
\bibliography{bib}


\begin{thebibliography}{62}


\ifx \showCODEN    \undefined \def \showCODEN     #1{\unskip}     \fi
\ifx \showDOI      \undefined \def \showDOI       #1{#1}\fi
\ifx \showISBNx    \undefined \def \showISBNx     #1{\unskip}     \fi
\ifx \showISBNxiii \undefined \def \showISBNxiii  #1{\unskip}     \fi
\ifx \showISSN     \undefined \def \showISSN      #1{\unskip}     \fi
\ifx \showLCCN     \undefined \def \showLCCN      #1{\unskip}     \fi
\ifx \shownote     \undefined \def \shownote      #1{#1}          \fi
\ifx \showarticletitle \undefined \def \showarticletitle #1{#1}   \fi
\ifx \showURL      \undefined \def \showURL       {\relax}        \fi
\providecommand\bibfield[2]{#2}
\providecommand\bibinfo[2]{#2}
\providecommand\natexlab[1]{#1}
\providecommand\showeprint[2][]{arXiv:#2}

\bibitem[\protect\citeauthoryear{Bochkovskiy, Wang, and Liao}{Bochkovskiy
  et~al\mbox{.}}{2020}]%
        {bochkovskiy2020yolov4}
\bibfield{author}{\bibinfo{person}{Alexey Bochkovskiy},
  \bibinfo{person}{Chien-Yao Wang}, {and} \bibinfo{person}{Hong-Yuan~Mark
  Liao}.} \bibinfo{year}{2020}\natexlab{}.
\newblock \showarticletitle{YOLOv4: Optimal speed and accuracy of object
  detection}.
\newblock \bibinfo{journal}{\emph{arXiv preprint arXiv:2004.10934}}
  (\bibinfo{year}{2020}).
\newblock


\bibitem[\protect\citeauthoryear{Cai and Vasconcelos}{Cai and
  Vasconcelos}{2018}]%
        {cai2018cascade}
\bibfield{author}{\bibinfo{person}{Zhaowei Cai} {and} \bibinfo{person}{Nuno
  Vasconcelos}.} \bibinfo{year}{2018}\natexlab{}.
\newblock \showarticletitle{{Cascade RCNN: Delving into high quality object
  detection}}. In \bibinfo{booktitle}{\emph{Proceedings of the IEEE/CVF
  Conference on Computer Vision and Pattern Recognition}}.
\newblock


\bibitem[\protect\citeauthoryear{Chen, Wang, Pang, Cao, Xiong, Li, Sun, Feng,
  Liu, Xu, et~al\mbox{.}}{Chen et~al\mbox{.}}{2019b}]%
        {chen2019mmdetection}
\bibfield{author}{\bibinfo{person}{Kai Chen}, \bibinfo{person}{Jiaqi Wang},
  \bibinfo{person}{Jiangmiao Pang}, \bibinfo{person}{Yuhang Cao},
  \bibinfo{person}{Yu Xiong}, \bibinfo{person}{Xiaoxiao Li},
  \bibinfo{person}{Shuyang Sun}, \bibinfo{person}{Wansen Feng},
  \bibinfo{person}{Ziwei Liu}, \bibinfo{person}{Jiarui Xu}, {et~al\mbox{.}}}
  \bibinfo{year}{2019}\natexlab{b}.
\newblock \showarticletitle{MMDetection: Open mmlab detection toolbox and
  benchmark}.
\newblock \bibinfo{journal}{\emph{arXiv preprint arXiv:1906.07155}}
  (\bibinfo{year}{2019}).
\newblock


\bibitem[\protect\citeauthoryear{Chen, Han, Wang, and Zhang}{Chen
  et~al\mbox{.}}{2019a}]%
        {chen2019revisiting}
\bibfield{author}{\bibinfo{person}{Yuntao Chen}, \bibinfo{person}{Chenxia Han},
  \bibinfo{person}{Naiyan Wang}, {and} \bibinfo{person}{Zhaoxiang Zhang}.}
  \bibinfo{year}{2019}\natexlab{a}.
\newblock \showarticletitle{Revisiting feature alignment for one-stage object
  detection}.
\newblock \bibinfo{journal}{\emph{arXiv preprint arXiv:1908.01570}}
  (\bibinfo{year}{2019}).
\newblock


\bibitem[\protect\citeauthoryear{Chen, Zhang, Cao, Wang, Lin, and Hu}{Chen
  et~al\mbox{.}}{2020}]%
        {chen2020reppoints}
\bibfield{author}{\bibinfo{person}{Yihong Chen}, \bibinfo{person}{Zheng Zhang},
  \bibinfo{person}{Yue Cao}, \bibinfo{person}{Liwei Wang},
  \bibinfo{person}{Stephen Lin}, {and} \bibinfo{person}{Han Hu}.}
  \bibinfo{year}{2020}\natexlab{}.
\newblock \showarticletitle{Reppoints v2: Verification meets regression for
  object detection}.
\newblock \bibinfo{journal}{\emph{Advances in Neural Information Processing
  Systems}}  \bibinfo{volume}{33} (\bibinfo{year}{2020}).
\newblock


\bibitem[\protect\citeauthoryear{Dai, Qi, Xiong, Li, Zhang, Hu, and Wei}{Dai
  et~al\mbox{.}}{2017}]%
        {dai2017deformable}
\bibfield{author}{\bibinfo{person}{Jifeng Dai}, \bibinfo{person}{Haozhi Qi},
  \bibinfo{person}{Yuwen Xiong}, \bibinfo{person}{Yi Li},
  \bibinfo{person}{Guodong Zhang}, \bibinfo{person}{Han Hu}, {and}
  \bibinfo{person}{Yichen Wei}.} \bibinfo{year}{2017}\natexlab{}.
\newblock \showarticletitle{Deformable convolutional networks}. In
  \bibinfo{booktitle}{\emph{Proceedings of the IEEE international conference on
  computer vision}}. \bibinfo{pages}{764--773}.
\newblock


\bibitem[\protect\citeauthoryear{Deng, Dong, Socher, Li, Li, and Fei-Fei}{Deng
  et~al\mbox{.}}{2009}]%
        {deng2009imagenet}
\bibfield{author}{\bibinfo{person}{Jia Deng}, \bibinfo{person}{Wei Dong},
  \bibinfo{person}{Richard Socher}, \bibinfo{person}{Li-Jia Li},
  \bibinfo{person}{Kai Li}, {and} \bibinfo{person}{Li Fei-Fei}.}
  \bibinfo{year}{2009}\natexlab{}.
\newblock \showarticletitle{Imagenet: A large-scale hierarchical image
  database}. In \bibinfo{booktitle}{\emph{Proceedings of the IEEE/CVF
  Conference on Computer Vision and Pattern Recognition}}.
  \bibinfo{pages}{248--255}.
\newblock


\bibitem[\protect\citeauthoryear{Deng, Guo, Zhou, Yu, Kotsia, and
  Zafeiriou}{Deng et~al\mbox{.}}{2019}]%
        {deng2019retinaface}
\bibfield{author}{\bibinfo{person}{Jiankang Deng}, \bibinfo{person}{Jia Guo},
  \bibinfo{person}{Yuxiang Zhou}, \bibinfo{person}{Jinke Yu},
  \bibinfo{person}{Irene Kotsia}, {and} \bibinfo{person}{Stefanos Zafeiriou}.}
  \bibinfo{year}{2019}\natexlab{}.
\newblock \showarticletitle{Retinaface: Single-stage dense face localisation in
  the wild}.
\newblock \bibinfo{journal}{\emph{arXiv preprint arXiv:1905.00641}}
  (\bibinfo{year}{2019}).
\newblock


\bibitem[\protect\citeauthoryear{Dong, Li, Liao, Wang, Ren, and Qian}{Dong
  et~al\mbox{.}}{2020}]%
        {dong2020centripetalnet}
\bibfield{author}{\bibinfo{person}{Zhiwei Dong}, \bibinfo{person}{Guoxuan Li},
  \bibinfo{person}{Yue Liao}, \bibinfo{person}{Fei Wang},
  \bibinfo{person}{Pengju Ren}, {and} \bibinfo{person}{Chen Qian}.}
  \bibinfo{year}{2020}\natexlab{}.
\newblock \showarticletitle{Centripetalnet: Pursuing high-quality keypoint
  pairs for object detection}. In \bibinfo{booktitle}{\emph{Proceedings of the
  IEEE/CVF Conference on Computer Vision and Pattern Recognition}}.
  \bibinfo{pages}{10519--10528}.
\newblock


\bibitem[\protect\citeauthoryear{Duan, Bai, Xie, Qi, Huang, and Tian}{Duan
  et~al\mbox{.}}{2019}]%
        {duan2019centernet}
\bibfield{author}{\bibinfo{person}{Kaiwen Duan}, \bibinfo{person}{Song Bai},
  \bibinfo{person}{Lingxi Xie}, \bibinfo{person}{Honggang Qi},
  \bibinfo{person}{Qingming Huang}, {and} \bibinfo{person}{Qi Tian}.}
  \bibinfo{year}{2019}\natexlab{}.
\newblock \showarticletitle{{Centernet: Keypoint triplets for object
  detection}}. In \bibinfo{booktitle}{\emph{Proceedings of the IEEE
  International Conference on Computer Vision}}.
\newblock


\bibitem[\protect\citeauthoryear{Gao, Cheng, Zhao, Zhang, Yang, and Torr}{Gao
  et~al\mbox{.}}{2021}]%
        {gao2019res2net}
\bibfield{author}{\bibinfo{person}{Shanghua Gao}, \bibinfo{person}{Ming-Ming
  Cheng}, \bibinfo{person}{Kai Zhao}, \bibinfo{person}{Xin-Yu Zhang},
  \bibinfo{person}{Ming-Hsuan Yang}, {and} \bibinfo{person}{Philip~HS Torr}.}
  \bibinfo{year}{2021}\natexlab{}.
\newblock \showarticletitle{Res2Net: A New Multi-Scale Backbone Architecture}.
\newblock \bibinfo{journal}{\emph{IEEE Transactions on Pattern Analysis and
  Machine Intelligence}} \bibinfo{volume}{43}, \bibinfo{number}{2}
  (\bibinfo{year}{2021}), \bibinfo{pages}{652--662}.
\newblock
\urldef\tempurl%
\url{https://doi.org/10.1109/TPAMI.2019.2938758}
\showDOI{\tempurl}


\bibitem[\protect\citeauthoryear{Girshick, Donahue, Darrell, and
  Malik}{Girshick et~al\mbox{.}}{2014}]%
        {girshick2014rich}
\bibfield{author}{\bibinfo{person}{Ross Girshick}, \bibinfo{person}{Jeff
  Donahue}, \bibinfo{person}{Trevor Darrell}, {and} \bibinfo{person}{Jitendra
  Malik}.} \bibinfo{year}{2014}\natexlab{}.
\newblock \showarticletitle{{Rich feature hierarchies for accurate object
  detection and semantic segmentation}}. In
  \bibinfo{booktitle}{\emph{Proceedings of the IEEE/CVF Conference on Computer
  Vision and Pattern Recognition}}.
\newblock


\bibitem[\protect\citeauthoryear{He, Zhang, Ren, and Sun}{He
  et~al\mbox{.}}{2016}]%
        {he2016deep}
\bibfield{author}{\bibinfo{person}{Kaiming He}, \bibinfo{person}{Xiangyu
  Zhang}, \bibinfo{person}{Shaoqing Ren}, {and} \bibinfo{person}{Jian Sun}.}
  \bibinfo{year}{2016}\natexlab{}.
\newblock \showarticletitle{{Deep residual learning for image recognition}}. In
  \bibinfo{booktitle}{\emph{Proceedings of the IEEE/CVF Conference on Computer
  Vision and Pattern Recognition}}.
\newblock


\bibitem[\protect\citeauthoryear{Huang, Yang, Deng, and Yu}{Huang
  et~al\mbox{.}}{2015}]%
        {huang2015densebox}
\bibfield{author}{\bibinfo{person}{Lichao Huang}, \bibinfo{person}{Yi Yang},
  \bibinfo{person}{Yafeng Deng}, {and} \bibinfo{person}{Yinan Yu}.}
  \bibinfo{year}{2015}\natexlab{}.
\newblock \showarticletitle{{DenseBox: Unifying landmark localization with end
  to end object detection}}.
\newblock \bibinfo{journal}{\emph{arXiv preprint arXiv:1509.04874}}
  (\bibinfo{year}{2015}).
\newblock


\bibitem[\protect\citeauthoryear{Jiang, Luo, Mao, Xiao, and Jiang}{Jiang
  et~al\mbox{.}}{2018}]%
        {jiang2018acquisition}
\bibfield{author}{\bibinfo{person}{Borui Jiang}, \bibinfo{person}{Ruixuan Luo},
  \bibinfo{person}{Jiayuan Mao}, \bibinfo{person}{Tete Xiao}, {and}
  \bibinfo{person}{Yuning Jiang}.} \bibinfo{year}{2018}\natexlab{}.
\newblock \showarticletitle{Acquisition of localization confidence for accurate
  object detection}. In \bibinfo{booktitle}{\emph{Proceedings of the European
  Conference on Computer Vision}}. \bibinfo{pages}{784--799}.
\newblock


\bibitem[\protect\citeauthoryear{Kim and Lee}{Kim and Lee}{2020}]%
        {kim2020probabilistic}
\bibfield{author}{\bibinfo{person}{Kang Kim} {and} \bibinfo{person}{Hee~Seok
  Lee}.} \bibinfo{year}{2020}\natexlab{}.
\newblock \showarticletitle{Probabilistic anchor assignment with iou prediction
  for object detection}.
\newblock \bibinfo{journal}{\emph{arXiv preprint arXiv:2007.08103}}
  (\bibinfo{year}{2020}).
\newblock


\bibitem[\protect\citeauthoryear{Law and Deng}{Law and Deng}{2018}]%
        {law2018cornernet}
\bibfield{author}{\bibinfo{person}{Hei Law} {and} \bibinfo{person}{Jia Deng}.}
  \bibinfo{year}{2018}\natexlab{}.
\newblock \showarticletitle{{CornerNet: Detecting objects as paired
  keypoints}}. In \bibinfo{booktitle}{\emph{Proceedings of the European
  Conference on Computer Vision}}.
\newblock


\bibitem[\protect\citeauthoryear{Li, Wang, Wang, Tai, Qian, Yang, Wang, Li, and
  Huang}{Li et~al\mbox{.}}{2019c}]%
        {li2019dsfd}
\bibfield{author}{\bibinfo{person}{Jian Li}, \bibinfo{person}{Yabiao Wang},
  \bibinfo{person}{Changan Wang}, \bibinfo{person}{Ying Tai},
  \bibinfo{person}{Jianjun Qian}, \bibinfo{person}{Jian Yang},
  \bibinfo{person}{Chengjie Wang}, \bibinfo{person}{Jilin Li}, {and}
  \bibinfo{person}{Feiyue Huang}.} \bibinfo{year}{2019}\natexlab{c}.
\newblock \showarticletitle{DSFD: dual shot face detector}. In
  \bibinfo{booktitle}{\emph{Proceedings of the IEEE/CVF Conference on Computer
  Vision and Pattern Recognition}}. \bibinfo{pages}{5060--5069}.
\newblock


\bibitem[\protect\citeauthoryear{Li, Wang, Wu, Chen, Hu, Li, Tang, and Yang}{Li
  et~al\mbox{.}}{2020}]%
        {li2020generalized}
\bibfield{author}{\bibinfo{person}{Xiang Li}, \bibinfo{person}{Wenhai Wang},
  \bibinfo{person}{Lijun Wu}, \bibinfo{person}{Shuo Chen},
  \bibinfo{person}{Xiaolin Hu}, \bibinfo{person}{Jun Li},
  \bibinfo{person}{Jinhui Tang}, {and} \bibinfo{person}{Jian Yang}.}
  \bibinfo{year}{2020}\natexlab{}.
\newblock \showarticletitle{Generalized Focal Loss: Learning Qualified and
  Distributed Bounding Boxes for Dense Object Detection}. In
  \bibinfo{booktitle}{\emph{Advances in Neural Information Processing
  Systems}}.
\newblock


\bibitem[\protect\citeauthoryear{Li, Chen, Wang, and Zhang}{Li
  et~al\mbox{.}}{2019a}]%
        {li2019scale}
\bibfield{author}{\bibinfo{person}{Yanghao Li}, \bibinfo{person}{Yuntao Chen},
  \bibinfo{person}{Naiyan Wang}, {and} \bibinfo{person}{Zhaoxiang Zhang}.}
  \bibinfo{year}{2019}\natexlab{a}.
\newblock \showarticletitle{{Scale-aware trident networks for object
  detection}}. In \bibinfo{booktitle}{\emph{Proceedings of the IEEE
  International Conference on Computer Vision}}.
\newblock


\bibitem[\protect\citeauthoryear{Li, Tang, Han, Liu, and He}{Li
  et~al\mbox{.}}{2019b}]%
        {li2019pyramidbox++}
\bibfield{author}{\bibinfo{person}{Zhihang Li}, \bibinfo{person}{Xu Tang},
  \bibinfo{person}{Junyu Han}, \bibinfo{person}{Jingtuo Liu}, {and}
  \bibinfo{person}{Ran He}.} \bibinfo{year}{2019}\natexlab{b}.
\newblock \showarticletitle{Pyramidbox++: High performance detector for finding
  tiny face}.
\newblock \bibinfo{journal}{\emph{arXiv preprint arXiv:1904.00386}}
  (\bibinfo{year}{2019}).
\newblock


\bibitem[\protect\citeauthoryear{Lin, Doll{\'a}r, Girshick, He, Hariharan, and
  Belongie}{Lin et~al\mbox{.}}{2017a}]%
        {lin2017feature}
\bibfield{author}{\bibinfo{person}{Tsung-Yi Lin}, \bibinfo{person}{Piotr
  Doll{\'a}r}, \bibinfo{person}{Ross Girshick}, \bibinfo{person}{Kaiming He},
  \bibinfo{person}{Bharath Hariharan}, {and} \bibinfo{person}{Serge Belongie}.}
  \bibinfo{year}{2017}\natexlab{a}.
\newblock \showarticletitle{{Feature pyramid networks for object detection}}.
  In \bibinfo{booktitle}{\emph{Proceedings of the IEEE/CVF Conference on
  Computer Vision and Pattern Recognition}}.
\newblock


\bibitem[\protect\citeauthoryear{Lin, Goyal, Girshick, He, and Doll{\'a}r}{Lin
  et~al\mbox{.}}{2017b}]%
        {lin2017focal}
\bibfield{author}{\bibinfo{person}{Tsung-Yi Lin}, \bibinfo{person}{Priya
  Goyal}, \bibinfo{person}{Ross Girshick}, \bibinfo{person}{Kaiming He}, {and}
  \bibinfo{person}{Piotr Doll{\'a}r}.} \bibinfo{year}{2017}\natexlab{b}.
\newblock \showarticletitle{{Focal loss for dense object detection}}. In
  \bibinfo{booktitle}{\emph{Proceedings of the IEEE International Conference on
  Computer Vision}}.
\newblock


\bibitem[\protect\citeauthoryear{Lin, Maire, Belongie, Hays, Perona, Ramanan,
  Doll{\'a}r, and Zitnick}{Lin et~al\mbox{.}}{2014}]%
        {lin2014microsoft}
\bibfield{author}{\bibinfo{person}{Tsung-Yi Lin}, \bibinfo{person}{Michael
  Maire}, \bibinfo{person}{Serge Belongie}, \bibinfo{person}{James Hays},
  \bibinfo{person}{Pietro Perona}, \bibinfo{person}{Deva Ramanan},
  \bibinfo{person}{Piotr Doll{\'a}r}, {and} \bibinfo{person}{C~Lawrence
  Zitnick}.} \bibinfo{year}{2014}\natexlab{}.
\newblock \showarticletitle{{Microsoft coco: Common objects in context}}. In
  \bibinfo{booktitle}{\emph{Proceedings of the European Conference on Computer
  Vision}}.
\newblock


\bibitem[\protect\citeauthoryear{Liu, Tang, Han, Liu, Rui, and Wu}{Liu
  et~al\mbox{.}}{2020}]%
        {Liu_2020_CVPR}
\bibfield{author}{\bibinfo{person}{Yang Liu}, \bibinfo{person}{Xu Tang},
  \bibinfo{person}{Junyu Han}, \bibinfo{person}{Jingtuo Liu},
  \bibinfo{person}{Dinger Rui}, {and} \bibinfo{person}{Xiang Wu}.}
  \bibinfo{year}{2020}\natexlab{}.
\newblock \showarticletitle{HAMBox: Delving Into Mining High-Quality Anchors on
  Face Detection}. In \bibinfo{booktitle}{\emph{IEEE/CVF Conference on Computer
  Vision and Pattern Recognition}}.
\newblock


\bibitem[\protect\citeauthoryear{Lu, Li, Yue, Li, and Yan}{Lu
  et~al\mbox{.}}{2019}]%
        {lu2019grid}
\bibfield{author}{\bibinfo{person}{Xin Lu}, \bibinfo{person}{Buyu Li},
  \bibinfo{person}{Yuxin Yue}, \bibinfo{person}{Quanquan Li}, {and}
  \bibinfo{person}{Junjie Yan}.} \bibinfo{year}{2019}\natexlab{}.
\newblock \showarticletitle{{Grid R-CNN}}. In
  \bibinfo{booktitle}{\emph{Proceedings of the IEEE/CVF Conference on Computer
  Vision and Pattern Recognition}}.
\newblock


\bibitem[\protect\citeauthoryear{Najibi, Samangouei, Chellappa, and
  Davis}{Najibi et~al\mbox{.}}{2017}]%
        {najibi2017ssh}
\bibfield{author}{\bibinfo{person}{Mahyar Najibi}, \bibinfo{person}{Pouya
  Samangouei}, \bibinfo{person}{Rama Chellappa}, {and} \bibinfo{person}{Larry~S
  Davis}.} \bibinfo{year}{2017}\natexlab{}.
\newblock \showarticletitle{Ssh: Single stage headless face detector}. In
  \bibinfo{booktitle}{\emph{Proceedings of the IEEE International Conference on
  Computer Vision}}. \bibinfo{pages}{4875--4884}.
\newblock


\bibitem[\protect\citeauthoryear{Najibi, Singh, and Davis}{Najibi
  et~al\mbox{.}}{2019}]%
        {najibi2019autofocus}
\bibfield{author}{\bibinfo{person}{Mahyar Najibi}, \bibinfo{person}{Bharat
  Singh}, {and} \bibinfo{person}{Larry~S Davis}.}
  \bibinfo{year}{2019}\natexlab{}.
\newblock \showarticletitle{{Autofocus: Efficient multi-scale inference}}. In
  \bibinfo{booktitle}{\emph{Proceedings of the IEEE International Conference on
  Computer Vision}}.
\newblock


\bibitem[\protect\citeauthoryear{Newell, Yang, and Deng}{Newell
  et~al\mbox{.}}{2016}]%
        {newell2016stacked}
\bibfield{author}{\bibinfo{person}{Alejandro Newell}, \bibinfo{person}{Kaiyu
  Yang}, {and} \bibinfo{person}{Jia Deng}.} \bibinfo{year}{2016}\natexlab{}.
\newblock \showarticletitle{Stacked hourglass networks for human pose
  estimation}. In \bibinfo{booktitle}{\emph{Proceedings of the European
  Conference on Computer Vision}}. Springer, \bibinfo{pages}{483--499}.
\newblock


\bibitem[\protect\citeauthoryear{Oksuz, Cam, Kalkan, and Akbas}{Oksuz
  et~al\mbox{.}}{2020}]%
        {oksuz2020imbalance}
\bibfield{author}{\bibinfo{person}{Kemal Oksuz}, \bibinfo{person}{Baris~Can
  Cam}, \bibinfo{person}{Sinan Kalkan}, {and} \bibinfo{person}{Emre Akbas}.}
  \bibinfo{year}{2020}\natexlab{}.
\newblock \showarticletitle{Imbalance problems in object detection: A review}.
\newblock \bibinfo{journal}{\emph{IEEE Transactions on Pattern Analysis and
  Machine Intelligence}} (\bibinfo{year}{2020}).
\newblock


\bibitem[\protect\citeauthoryear{Pang, Chen, Shi, Feng, Ouyang, and Lin}{Pang
  et~al\mbox{.}}{2019}]%
        {pang2019libra}
\bibfield{author}{\bibinfo{person}{Jiangmiao Pang}, \bibinfo{person}{Kai Chen},
  \bibinfo{person}{Jianping Shi}, \bibinfo{person}{Huajun Feng},
  \bibinfo{person}{Wanli Ouyang}, {and} \bibinfo{person}{Dahua Lin}.}
  \bibinfo{year}{2019}\natexlab{}.
\newblock \showarticletitle{{Libra r-cnn: Towards balanced learning for object
  detection}}. In \bibinfo{booktitle}{\emph{Proceedings of the IEEE/CVF
  Conference on Computer Vision and Pattern Recognition}}.
\newblock


\bibitem[\protect\citeauthoryear{Qiu, Ma, Li, Liu, and Sun}{Qiu
  et~al\mbox{.}}{2020}]%
        {qiu2020borderdet}
\bibfield{author}{\bibinfo{person}{Han Qiu}, \bibinfo{person}{Yuchen Ma},
  \bibinfo{person}{Zeming Li}, \bibinfo{person}{Songtao Liu}, {and}
  \bibinfo{person}{Jian Sun}.} \bibinfo{year}{2020}\natexlab{}.
\newblock \showarticletitle{{Borderdet: Border Feature for Dense Object
  Detection}}. In \bibinfo{booktitle}{\emph{Proceedings of the European
  Conference on Computer Vision}}.
\newblock


\bibitem[\protect\citeauthoryear{Ren, He, Girshick, and Sun}{Ren
  et~al\mbox{.}}{2015}]%
        {ren2015faster}
\bibfield{author}{\bibinfo{person}{Shaoqing Ren}, \bibinfo{person}{Kaiming He},
  \bibinfo{person}{Ross Girshick}, {and} \bibinfo{person}{Jian Sun}.}
  \bibinfo{year}{2015}\natexlab{}.
\newblock \showarticletitle{{Faster r-cnn: Towards real-time object detection
  with region proposal networks}}. In \bibinfo{booktitle}{\emph{Advances in
  Neural Information Processing Systems}}.
\newblock


\bibitem[\protect\citeauthoryear{Song, Liu, and Wang}{Song
  et~al\mbox{.}}{2020}]%
        {song2020revisiting}
\bibfield{author}{\bibinfo{person}{Guanglu Song}, \bibinfo{person}{Yu Liu},
  {and} \bibinfo{person}{Xiaogang Wang}.} \bibinfo{year}{2020}\natexlab{}.
\newblock \showarticletitle{Revisiting the sibling head in object detector}. In
  \bibinfo{booktitle}{\emph{Proceedings of the IEEE/CVF Conference on Computer
  Vision and Pattern Recognition}}. \bibinfo{pages}{11563--11572}.
\newblock


\bibitem[\protect\citeauthoryear{Sun, Jiang, Xie, Yuan, Wang, and Luo}{Sun
  et~al\mbox{.}}{2020}]%
        {sun2020onenet}
\bibfield{author}{\bibinfo{person}{Peize Sun}, \bibinfo{person}{Yi Jiang},
  \bibinfo{person}{Enze Xie}, \bibinfo{person}{Zehuan Yuan},
  \bibinfo{person}{Changhu Wang}, {and} \bibinfo{person}{Ping Luo}.}
  \bibinfo{year}{2020}\natexlab{}.
\newblock \showarticletitle{OneNet: Towards End-to-End One-Stage Object
  Detection}.
\newblock \bibinfo{journal}{\emph{arXiv preprint arXiv:2012.05780}}
  (\bibinfo{year}{2020}).
\newblock


\bibitem[\protect\citeauthoryear{Tan, Lu, Zhang, Yin, and Li}{Tan
  et~al\mbox{.}}{2020a}]%
        {tan2020equalizationv2}
\bibfield{author}{\bibinfo{person}{Jingru Tan}, \bibinfo{person}{Xin Lu},
  \bibinfo{person}{Gang Zhang}, \bibinfo{person}{Changqing Yin}, {and}
  \bibinfo{person}{Quanquan Li}.} \bibinfo{year}{2020}\natexlab{a}.
\newblock \showarticletitle{Equalization Loss v2: A New Gradient Balance
  Approach for Long-tailed Object Detection}.
\newblock \bibinfo{journal}{\emph{arXiv preprint arXiv:2012.08548}}
  (\bibinfo{year}{2020}).
\newblock


\bibitem[\protect\citeauthoryear{Tan, Wang, Li, Li, Ouyang, Yin, and Yan}{Tan
  et~al\mbox{.}}{2020b}]%
        {tan2020equalization}
\bibfield{author}{\bibinfo{person}{Jingru Tan}, \bibinfo{person}{Changbao
  Wang}, \bibinfo{person}{Buyu Li}, \bibinfo{person}{Quanquan Li},
  \bibinfo{person}{Wanli Ouyang}, \bibinfo{person}{Changqing Yin}, {and}
  \bibinfo{person}{Junjie Yan}.} \bibinfo{year}{2020}\natexlab{b}.
\newblock \showarticletitle{Equalization loss for long-tailed object
  recognition}. In \bibinfo{booktitle}{\emph{Proceedings of the IEEE/CVF
  Conference on Computer Vision and Pattern Recognition}}.
  \bibinfo{pages}{11662--11671}.
\newblock


\bibitem[\protect\citeauthoryear{Tang, Du, He, and Liu}{Tang
  et~al\mbox{.}}{2018}]%
        {tang2018pyramidbox}
\bibfield{author}{\bibinfo{person}{Xu Tang}, \bibinfo{person}{Daniel~K Du},
  \bibinfo{person}{Zeqiang He}, {and} \bibinfo{person}{Jingtuo Liu}.}
  \bibinfo{year}{2018}\natexlab{}.
\newblock \showarticletitle{Pyramidbox: A context-assisted single shot face
  detector}. In \bibinfo{booktitle}{\emph{Proceedings of the European
  Conference on Computer Vision}}. \bibinfo{pages}{797--813}.
\newblock


\bibitem[\protect\citeauthoryear{Tian, Shen, Chen, and He}{Tian
  et~al\mbox{.}}{2019}]%
        {tian2019fcos}
\bibfield{author}{\bibinfo{person}{Zhi Tian}, \bibinfo{person}{Chunhua Shen},
  \bibinfo{person}{Hao Chen}, {and} \bibinfo{person}{Tong He}.}
  \bibinfo{year}{2019}\natexlab{}.
\newblock \showarticletitle{{Fcos: Fully convolutional one-stage object
  detection}}. In \bibinfo{booktitle}{\emph{Proceedings of the IEEE
  International Conference on Computer Vision}}.
\newblock


\bibitem[\protect\citeauthoryear{Vu, Jang, Pham, and Yoo}{Vu
  et~al\mbox{.}}{2019}]%
        {NEURIPS2019_d554f7bb}
\bibfield{author}{\bibinfo{person}{Thang Vu}, \bibinfo{person}{Hyunjun Jang},
  \bibinfo{person}{Trung~X. Pham}, {and} \bibinfo{person}{Chang Yoo}.}
  \bibinfo{year}{2019}\natexlab{}.
\newblock \showarticletitle{Cascade RPN: Delving into High-Quality Region
  Proposal Network with Adaptive Convolution}. In
  \bibinfo{booktitle}{\emph{Advances in Neural Information Processing
  Systems}}, Vol.~\bibinfo{volume}{32}. \bibinfo{publisher}{Curran Associates,
  Inc.}
\newblock


\bibitem[\protect\citeauthoryear{Wang, Bochkovskiy, and Liao}{Wang
  et~al\mbox{.}}{2020a}]%
        {wang2020scaled}
\bibfield{author}{\bibinfo{person}{Chien-Yao Wang}, \bibinfo{person}{Alexey
  Bochkovskiy}, {and} \bibinfo{person}{Hong-Yuan~Mark Liao}.}
  \bibinfo{year}{2020}\natexlab{a}.
\newblock \showarticletitle{Scaled-YOLOv4: Scaling Cross Stage Partial
  Network}.
\newblock \bibinfo{journal}{\emph{arXiv preprint arXiv:2011.08036}}
  (\bibinfo{year}{2020}).
\newblock


\bibitem[\protect\citeauthoryear{Wang, Chen, Yang, Loy, and Lin}{Wang
  et~al\mbox{.}}{2019}]%
        {wang2019region}
\bibfield{author}{\bibinfo{person}{Jiaqi Wang}, \bibinfo{person}{Kai Chen},
  \bibinfo{person}{Shuo Yang}, \bibinfo{person}{Chen~Change Loy}, {and}
  \bibinfo{person}{Dahua Lin}.} \bibinfo{year}{2019}\natexlab{}.
\newblock \showarticletitle{{Region proposal by guided anchoring}}. In
  \bibinfo{booktitle}{\emph{Proceedings of the IEEE/CVF Conference on Computer
  Vision and Pattern Recognition}}.
\newblock


\bibitem[\protect\citeauthoryear{Wang, Song, Li, Sun, Sun, and Zheng}{Wang
  et~al\mbox{.}}{2020b}]%
        {wang2020end}
\bibfield{author}{\bibinfo{person}{Jianfeng Wang}, \bibinfo{person}{Lin Song},
  \bibinfo{person}{Zeming Li}, \bibinfo{person}{Hongbin Sun},
  \bibinfo{person}{Jian Sun}, {and} \bibinfo{person}{Nanning Zheng}.}
  \bibinfo{year}{2020}\natexlab{b}.
\newblock \showarticletitle{End-to-end object detection with fully
  convolutional network}.
\newblock \bibinfo{journal}{\emph{arXiv preprint arXiv:2012.03544}}
  (\bibinfo{year}{2020}).
\newblock


\bibitem[\protect\citeauthoryear{Wang, Zhang, Cao, Chen, Pang, Gong, Shi, Loy,
  and Lin}{Wang et~al\mbox{.}}{2020c}]%
        {wang2019side}
\bibfield{author}{\bibinfo{person}{Jiaqi Wang}, \bibinfo{person}{Wenwei Zhang},
  \bibinfo{person}{Yuhang Cao}, \bibinfo{person}{Kai Chen},
  \bibinfo{person}{Jiangmiao Pang}, \bibinfo{person}{Tao Gong},
  \bibinfo{person}{Jianping Shi}, \bibinfo{person}{Chen~Change Loy}, {and}
  \bibinfo{person}{Dahua Lin}.} \bibinfo{year}{2020}\natexlab{c}.
\newblock \showarticletitle{{Side-aware boundary localization for more precise
  object detection}}.
\newblock \bibinfo{journal}{\emph{Proceedings of the European Conference on
  Computer Vision}} (\bibinfo{year}{2020}).
\newblock


\bibitem[\protect\citeauthoryear{Wang and Zhang}{Wang and Zhang}{2020}]%
        {10.1145/3394171.3413691}
\bibfield{author}{\bibinfo{person}{Keyang Wang} {and} \bibinfo{person}{Lei
  Zhang}.} \bibinfo{year}{2020}\natexlab{}.
\newblock \showarticletitle{Single-Shot Two-Pronged Detector with Rectified IoU
  Loss}. In \bibinfo{booktitle}{\emph{Proceedings of the 28th ACM International
  Conference on Multimedia}} (Seattle, WA, USA) \emph{(\bibinfo{series}{MM
  '20})}. \bibinfo{address}{New York, NY, USA}, \bibinfo{pages}{1311–1319}.
\newblock


\bibitem[\protect\citeauthoryear{Wu, Song, Wang, Zhang, and Yuan}{Wu
  et~al\mbox{.}}{2020b}]%
        {wu2020forest}
\bibfield{author}{\bibinfo{person}{Jialian Wu}, \bibinfo{person}{Liangchen
  Song}, \bibinfo{person}{Tiancai Wang}, \bibinfo{person}{Qian Zhang}, {and}
  \bibinfo{person}{Junsong Yuan}.} \bibinfo{year}{2020}\natexlab{b}.
\newblock \showarticletitle{Forest r-cnn: Large-vocabulary long-tailed object
  detection and instance segmentation}. In
  \bibinfo{booktitle}{\emph{Proceedings of the 28th ACM International
  Conference on Multimedia}}. \bibinfo{pages}{1570--1578}.
\newblock


\bibitem[\protect\citeauthoryear{Wu, Chen, Yuan, Liu, Wang, Li, and Fu}{Wu
  et~al\mbox{.}}{2020a}]%
        {wu2020rethinking}
\bibfield{author}{\bibinfo{person}{Yue Wu}, \bibinfo{person}{Yinpeng Chen},
  \bibinfo{person}{Lu Yuan}, \bibinfo{person}{Zicheng Liu},
  \bibinfo{person}{Lijuan Wang}, \bibinfo{person}{Hongzhi Li}, {and}
  \bibinfo{person}{Yun Fu}.} \bibinfo{year}{2020}\natexlab{a}.
\newblock \showarticletitle{{Rethinking Classification and Localization for
  Object Detection}}. In \bibinfo{booktitle}{\emph{Proceedings of the IEEE/CVF
  Conference on Computer Vision and Pattern Recognition}}.
\newblock


\bibitem[\protect\citeauthoryear{Xie, Girshick, Doll{\'a}r, Tu, and He}{Xie
  et~al\mbox{.}}{2017}]%
        {xie2017aggregated}
\bibfield{author}{\bibinfo{person}{Saining Xie}, \bibinfo{person}{Ross
  Girshick}, \bibinfo{person}{Piotr Doll{\'a}r}, \bibinfo{person}{Zhuowen Tu},
  {and} \bibinfo{person}{Kaiming He}.} \bibinfo{year}{2017}\natexlab{}.
\newblock \showarticletitle{{Aggregated Residual Transformations for Deep
  Neural Networks}}. In \bibinfo{booktitle}{\emph{Proceedings of the IEEE/CVF
  Conference on Computer Vision and Pattern Recognition}}.
\newblock


\bibitem[\protect\citeauthoryear{Yang, Huang, and Wang}{Yang
  et~al\mbox{.}}{2021}]%
        {yang2021querydet}
\bibfield{author}{\bibinfo{person}{Chenhongyi Yang}, \bibinfo{person}{Zehao
  Huang}, {and} \bibinfo{person}{Naiyan Wang}.}
  \bibinfo{year}{2021}\natexlab{}.
\newblock \showarticletitle{QueryDet: Cascaded Sparse Query for Accelerating
  High-Resolution Small Object Detection}.
\newblock \bibinfo{journal}{\emph{arXiv preprint arXiv:2103.09136}}
  (\bibinfo{year}{2021}).
\newblock


\bibitem[\protect\citeauthoryear{Yang, Luo, Loy, and Tang}{Yang
  et~al\mbox{.}}{2016}]%
        {yang2016wider}
\bibfield{author}{\bibinfo{person}{Shuo Yang}, \bibinfo{person}{Ping Luo},
  \bibinfo{person}{Chen-Change Loy}, {and} \bibinfo{person}{Xiaoou Tang}.}
  \bibinfo{year}{2016}\natexlab{}.
\newblock \showarticletitle{Wider face: A face detection benchmark}. In
  \bibinfo{booktitle}{\emph{Proceedings of the IEEE Conference on Computer
  Vision and Pattern Recognition}}. \bibinfo{pages}{5525--5533}.
\newblock


\bibitem[\protect\citeauthoryear{Yang, Liu, Hu, Wang, and Lin}{Yang
  et~al\mbox{.}}{2019}]%
        {yang2019reppoints}
\bibfield{author}{\bibinfo{person}{Ze Yang}, \bibinfo{person}{Shaohui Liu},
  \bibinfo{person}{Han Hu}, \bibinfo{person}{Liwei Wang}, {and}
  \bibinfo{person}{Stephen Lin}.} \bibinfo{year}{2019}\natexlab{}.
\newblock \showarticletitle{{Reppoints: Point set representation for object
  detection}}. In \bibinfo{booktitle}{\emph{Proceedings of the IEEE
  International Conference on Computer Vision}}.
\newblock


\bibitem[\protect\citeauthoryear{Zhang, Li, Wang, Tai, Wang, Li, Huang, Xia,
  Pei, and Ji}{Zhang et~al\mbox{.}}{2020c}]%
        {zhang2020asfd}
\bibfield{author}{\bibinfo{person}{Bin Zhang}, \bibinfo{person}{Jian Li},
  \bibinfo{person}{Yabiao Wang}, \bibinfo{person}{Ying Tai},
  \bibinfo{person}{Chengjie Wang}, \bibinfo{person}{Jilin Li},
  \bibinfo{person}{Feiyue Huang}, \bibinfo{person}{Yili Xia},
  \bibinfo{person}{Wenjiang Pei}, {and} \bibinfo{person}{Rongrong Ji}.}
  \bibinfo{year}{2020}\natexlab{c}.
\newblock \showarticletitle{ASFD: Automatic and Scalable Face Detector}.
\newblock \bibinfo{journal}{\emph{arXiv preprint arXiv:2003.11228}}
  (\bibinfo{year}{2020}).
\newblock


\bibitem[\protect\citeauthoryear{Zhang, Fan, Ai, Song, Qin, and Wu}{Zhang
  et~al\mbox{.}}{2019a}]%
        {zhang2019accurate}
\bibfield{author}{\bibinfo{person}{Faen Zhang}, \bibinfo{person}{Xinyu Fan},
  \bibinfo{person}{Guo Ai}, \bibinfo{person}{Jianfei Song},
  \bibinfo{person}{Yongqiang Qin}, {and} \bibinfo{person}{Jiahong Wu}.}
  \bibinfo{year}{2019}\natexlab{a}.
\newblock \showarticletitle{Accurate face detection for high performance}.
\newblock \bibinfo{journal}{\emph{arXiv preprint arXiv:1905.01585}}
  (\bibinfo{year}{2019}).
\newblock


\bibitem[\protect\citeauthoryear{Zhang, Chi, Lei, and Li}{Zhang
  et~al\mbox{.}}{2020a}]%
        {zhang2020refineface}
\bibfield{author}{\bibinfo{person}{Shifeng Zhang}, \bibinfo{person}{Cheng Chi},
  \bibinfo{person}{Zhen Lei}, {and} \bibinfo{person}{Stan~Z Li}.}
  \bibinfo{year}{2020}\natexlab{a}.
\newblock \showarticletitle{Refineface: Refinement neural network for high
  performance face detection}.
\newblock \bibinfo{journal}{\emph{IEEE Transactions on Pattern Analysis and
  Machine Intelligence}} (\bibinfo{year}{2020}).
\newblock


\bibitem[\protect\citeauthoryear{Zhang, Chi, Yao, Lei, and Li}{Zhang
  et~al\mbox{.}}{2020b}]%
        {zhang2020bridging}
\bibfield{author}{\bibinfo{person}{Shifeng Zhang}, \bibinfo{person}{Cheng Chi},
  \bibinfo{person}{Yongqiang Yao}, \bibinfo{person}{Zhen Lei}, {and}
  \bibinfo{person}{Stan~Z Li}.} \bibinfo{year}{2020}\natexlab{b}.
\newblock \showarticletitle{{Bridging the gap between anchor-based and
  anchor-free detection via adaptive training sample selection}}. In
  \bibinfo{booktitle}{\emph{Proceedings of the IEEE/CVF Conference on Computer
  Vision and Pattern Recognition}}.
\newblock


\bibitem[\protect\citeauthoryear{Zhang, Wan, Liu, Ji, and Ye}{Zhang
  et~al\mbox{.}}{2019b}]%
        {NEURIPS2019_43ec517d}
\bibfield{author}{\bibinfo{person}{Xiaosong Zhang}, \bibinfo{person}{Fang Wan},
  \bibinfo{person}{Chang Liu}, \bibinfo{person}{Rongrong Ji}, {and}
  \bibinfo{person}{Qixiang Ye}.} \bibinfo{year}{2019}\natexlab{b}.
\newblock \showarticletitle{FreeAnchor: Learning to Match Anchors for Visual
  Object Detection}. In \bibinfo{booktitle}{\emph{Advances in Neural
  Information Processing Systems}}, Vol.~\bibinfo{volume}{32}.
\newblock


\bibitem[\protect\citeauthoryear{Zhang, Wan, Liu, Ji, and Ye}{Zhang
  et~al\mbox{.}}{2021}]%
        {zhang2021learning}
\bibfield{author}{\bibinfo{person}{Xiaosong Zhang}, \bibinfo{person}{Fang Wan},
  \bibinfo{person}{Chang Liu}, \bibinfo{person}{Xiangyang Ji}, {and}
  \bibinfo{person}{Qixiang Ye}.} \bibinfo{year}{2021}\natexlab{}.
\newblock \showarticletitle{Learning to match anchors for visual object
  detection}.
\newblock \bibinfo{journal}{\emph{IEEE Transactions on Pattern Analysis and
  Machine Intelligence}} (\bibinfo{year}{2021}).
\newblock


\bibitem[\protect\citeauthoryear{Zheng, Wang, Liu, Li, Ye, and Ren}{Zheng
  et~al\mbox{.}}{2020}]%
        {zheng2020distance}
\bibfield{author}{\bibinfo{person}{Zhaohui Zheng}, \bibinfo{person}{Ping Wang},
  \bibinfo{person}{Wei Liu}, \bibinfo{person}{Jinze Li},
  \bibinfo{person}{Rongguang Ye}, {and} \bibinfo{person}{Dongwei Ren}.}
  \bibinfo{year}{2020}\natexlab{}.
\newblock \showarticletitle{Distance-IoU loss: Faster and better learning for
  bounding box regression}. In \bibinfo{booktitle}{\emph{Proceedings of the
  AAAI Conference on Artificial Intelligence}}, Vol.~\bibinfo{volume}{34}.
  \bibinfo{pages}{12993--13000}.
\newblock


\bibitem[\protect\citeauthoryear{Zhou, Sun, Bau, and Torralba}{Zhou
  et~al\mbox{.}}{2018}]%
        {zhou2018interpretable}
\bibfield{author}{\bibinfo{person}{Bolei Zhou}, \bibinfo{person}{Yiyou Sun},
  \bibinfo{person}{David Bau}, {and} \bibinfo{person}{Antonio Torralba}.}
  \bibinfo{year}{2018}\natexlab{}.
\newblock \showarticletitle{Interpretable basis decomposition for visual
  explanation}. In \bibinfo{booktitle}{\emph{Proceedings of the European
  Conference on Computer Vision}}. \bibinfo{pages}{119--134}.
\newblock


\bibitem[\protect\citeauthoryear{Zhu, Chen, Shen, and Savvides}{Zhu
  et~al\mbox{.}}{2020b}]%
        {zhu2019soft}
\bibfield{author}{\bibinfo{person}{Chenchen Zhu}, \bibinfo{person}{Fangyi
  Chen}, \bibinfo{person}{Zhiqiang Shen}, {and} \bibinfo{person}{Marios
  Savvides}.} \bibinfo{year}{2020}\natexlab{b}.
\newblock \showarticletitle{{Soft anchor-point object detection}}.
\newblock \bibinfo{journal}{\emph{Proceedings of the European Conference on
  Computer Vision}} (\bibinfo{year}{2020}).
\newblock


\bibitem[\protect\citeauthoryear{Zhu, He, and Savvides}{Zhu
  et~al\mbox{.}}{2019}]%
        {zhu2019feature}
\bibfield{author}{\bibinfo{person}{Chenchen Zhu}, \bibinfo{person}{Yihui He},
  {and} \bibinfo{person}{Marios Savvides}.} \bibinfo{year}{2019}\natexlab{}.
\newblock \showarticletitle{{Feature selective anchor-free module for
  single-shot object detection}}. In \bibinfo{booktitle}{\emph{Proceedings of
  the IEEE/CVF Conference on Computer Vision and Pattern Recognition}}.
\newblock


\bibitem[\protect\citeauthoryear{Zhu, Cai, Zhang, Wang, and Xiong}{Zhu
  et~al\mbox{.}}{2020a}]%
        {zhu2020tinaface}
\bibfield{author}{\bibinfo{person}{Yanjia Zhu}, \bibinfo{person}{Hongxiang
  Cai}, \bibinfo{person}{Shuhan Zhang}, \bibinfo{person}{Chenhao Wang}, {and}
  \bibinfo{person}{Yichao Xiong}.} \bibinfo{year}{2020}\natexlab{a}.
\newblock \showarticletitle{TinaFace: Strong but Simple Baseline for Face
  Detection}.
\newblock \bibinfo{journal}{\emph{arXiv preprint arXiv:2011.13183}}
  (\bibinfo{year}{2020}).
\newblock


\end{thebibliography}
}

\end{document}